# Symmetric and antisymmetric properties of solutions to kernel-based machine learning problems

**Giorgio Gnecco**                                                          giorgio.gnecco@imtlucca.it

*IMT - School for Advanced Studies*
*Piazza S. Francesco, 19 - 55110 Lucca, Italy*

## Abstract

A particularly interesting instance of supervised learning with kernels is when each training example is associated with two objects, as in pairwise classification (Brunner et al., 2012), and in supervised learning of preference relations (Herbrich et al., 1998). In these cases, one may want to embed additional prior knowledge into the optimization problem associated with the training of the learning machine, modeled, respectively, by the symmetry of its optimal solution with respect to an exchange of order between the two objects, and by its antisymmetry. Extending the approach proposed in (Brunner et al., 2012) (where the only symmetric case was considered), we show, focusing on support vector binary classification, how such embedding is possible through the choice of a suitable pairwise kernel, which takes as inputs the individual feature vectors and also the group feature vectors associated with the two objects. We also prove that the symmetry/antisymmetry constraints still hold when considering the sequence of suboptimal solutions generated by one version of the Sequential Minimal Optimization (SMO) algorithm, and we present numerical results supporting the theoretical findings. We conclude discussing extensions of the main results to support vector regression, to transductive support vector machines, and to several kinds of graph kernels, including diffusion kernels.

## 1. Introduction

In recent years, optimization theory has found several applications in machine learning problems: often, these are formulated as optimization problems in which, given a finite set of empirical data (e.g., a finite set of feature vectors, where each feature may represent the output of a measurement process), it is required to find a parameter vector associated with a model of such data, which is optimal according to a suitable performance index. In the context of the so-called kernel methods (Shawe-Taylor and Cristianini, 2004), examples of such problems are supervised (Cucker and Smale, 2002), unsupervised (von Luxburg, 2007), and semi-supervised (Belkin and Niyogi, 2006; Melacci and Belkin, 2011) learning, and identification of models of dynamical systems (Pillonetto et al., 2014). The performance index is typically composed of the sum of an empirical term (which measures the fitness of the model associated with a specific parameter vector in explaining the empirical data), and a regularization term, whose goal is to endow the optimal model with the capability of generalizing its predictions to new data, not used during the training of the model.

Additional a-priori/a-posteriori knowledge can be inserted into the optimization problem above, either requiring the satisfaction of given constraints, or penalizing in a suitable way their violation: see, e.g., the works (Gnecco et al., 2013, 2014, 2015a,b), which describe a novel approach to machine learning called learning from constraints and based, respectively, on hard and soft formulations of such constraints. Indeed, several experiments, such as those reported in (Diligenti et al., 2012, Section 6), show that taking into account constraints representing additional knowledge valid for a specific machine learning problem can improve the generalization capability of the learned model. Another possible way to deal with constraints on the optimal solution of a machine learning problem is to check a-posteriori if such constraints are satisfied by its optimal solution, e.g., as a consequence of necessary optimality conditions for the specific optimization problem modeling learning. This can be also considered as a particular case of learning from constraints, and is the approach followed in this work.

**Contributions.** Within this framework, the goal of this paper consists in investigating a particular instance of machine learning problems belonging to the learning from constraints framework. More precisely, we investigate the case of constraints associated, respectively, with symmetry and antisymmetry of the optimal solution (with respect to a suitable transformation of the feature vectors). Such constraints are of interest, e.g., for supervised learning problems in which each supervised example is associated with two objects, as in pairwise classification (e.g., in recognizing whether two face images belong to the same person or not) (Brunner et al., 2012), and in supervised learning of preference relations (i.e., in learning an order among objects through supervised pairs of examples) (Herbrich et al., 1998). Indeed, in the first case, the label does not depend on the order of the objects (e.g., two face images) in the pair, hence also the classification produced by the trained learning machine should not depend on such an order, and a symmetry constraint is needed. In the second case, instead, the label changes sign by changing the order of the objects (e.g., two football teams that one wants to compare in performance), hence also the output of the trained learning machine should change sign, and an antisymmetry constraint is required. After introducing in Section 2 the kind of machine learning problems considered in the paper, and some basic notation in Section 3, the paper starts with the investigation of two methods to impose the satisfaction of symmetry/antisymmetry constraints, in the context of kernel methods, focusing on binary classification: the use of either symmetric or antisymmetric training sets (together with the choice of a suitable kernel, see Section 4.1), and the use of a reduced training set (together with the choice of another suitable kernel, see Section 4.2). In both cases, the elements of the training sets are ordered pairs of objects, together with their labels. Since such ordered pairs can be associated in an one-to-one way with directed arcs on a graph, the machine learning problems investigated in the paper have also application on the binary classification of directed arcs on graphs, using, for each directed arc, a set of features associated with its nodes. In this case, imposing the symmetry/antisymmetry properties above can be interpreted as assigning, respectively, the same labels and classifications or opposite labels and classifications to the two directed arcs (with opposite orientations) connecting the same pair of nodes, for each of such pairs. The starting point of this part of our investigation is the work (Brunner et al., 2012), in which the only case of symmetry constraints was studied. It is worth mentioning that, according to (Brunner et al., 2012), that work is the first one in which the symmetry of the optimal solution to the training problem of an $l_1$-soft margin binary Support Vector Machine (SVM) classifier with a symmetric training set has been rigorously proved, for a quite general class of kernels. Taking inspiration from the invariance framework of (Király et al., 2014), we specialize this result to the presence of both individual features and two kinds of group features, then we extend it to the case of an antisymmetric training set, showing the antisymmetry of the optimal solution. This part of the investigation is mainly based on duality theory in optimization (Bertsekas, 1999) and on representer theorems for SVMs (Shawe-Taylor and Cristianini, 2004). We also employ existence and uniqueness results from the theory of SVMs (Burges and Crisp, 2000), and we introduce a kernel (called in the paper skew-balanced kernel), related specifically with the antisymmetry constraint. Additionally, in Section 5, we investigate how to impose the symmetry and antisymmetry constraints when looking for suboptimal solutions to the optimization problems investigated in the paper, analyzing the behavior of a specific algorithm proposed in the literature (Platt, 1999) from this viewpoint (i.e., examining if its steps preserve or not the symmetry and antisymmetry properties). Up to our knowledge, this kind of investigation is completely novel in the literature about kernel methods. In Section 6, we provide numerical results, which are in accordance with the ones obtained theoretically. Such results also demonstrate the practical advantage of using kernels enforcing symmetry/antisymmetry of the optimal solution to the specific machine learning problem (when the model generating the data labels also satisfies one of such properties), as compared with kernels not enforcing symmetry/antisymmetry. In Section 7, we also show how our analysis can be extended to other kernel methods, namely, to support vector regression (Section 7.1) and to transductive SVMs (Section 7.2), whose application in (Dardard et al., 2016) inspired part of the theoretical investigations of the present work, providing numerical results that showed the occurrence of the antisymmetry property, for a specific choice of the kernel (the linear kernel), and to which we refer for real-world examples of individual and group features in a motion-capture context. No such theoretical analysis was performed in (Dardard et al., 2016), where the antisymmetry property was only observed a-posteriori, after training the learning machine. Finally, we outline some extensions of our analysis to the problem of arc classification on graphs, a problem for which the application of pairwise kernels was not considered in (Brunner et al., 2012). For this specific case, in Sections 7.3 and 7.4, we also define suitable graph kernels related to features associated with several pairs of nodes in the graph (including diffusion kernels on an auxiliary graph), again with the goal to impose either symmetry or antisymmetry to the optimal solution of the associated machine learning problem. More specifically, we show how to construct graph kernels inheriting the properties of the kernels considered in the first part of the paper, as such properties are useful to impose symmetry/antisymmetry constraints.

**Related works.** Besides the work (Brunner et al., 2012) mentioned above, a related paper is (Király et al., 2014), which provides the general structure of kernels incorporating prior knowledge represented by algebraic invariances. Although in principle the theory developed therein (particularly, its Theorem 3 (i)) could be used also to generate kernels satisfying the specific symmetry property considered in the present paper (but not to generate kernels satisfying, instead, the antisymmetry property), the discussion of permutation invariances needed for that purpose is only briefly detailed in Section 2.5.6 of that paper. Moreover, even though it is related to the $G$-invariant kernels of (Király et al., 2014), the order-invariant kernel considered in this paper is not a particular case of the former, since, in the context of the present paper, the symmetry property for a symmetric training set is not associated only with the choice of the kernel, but also with the existence of optimal symmetric dual variables. Another related work is (Vedaldi et al., 2011), where a novel approach, based on a convex relaxation of a more general nonconvex optimization problem, is proposed to incorporate invariance and equivariance in the SVM training problem. However, such an approach does not guarantee these properties for all possible choices of its parameters, as demonstrated by the numerical results presented therein. Likewise in the present paper, symmetric and antisymmetric pairwise kernels have been also considered in (Waegeman et al., 2012), which presents sufficient conditions under which the use of a symmetric (respectively, antisymmetric) pairwise kernel makes it possible to approximate arbitrarily well any symmetric (respectively, antisymmetric) continuous function on a compact set. However, neither the extensions of the results of (Brunner et al., 2012) mentioned above are provided in (Waegeman et al., 2012), nor an investigation of the issue of symmetry/antisymmetry preservation of suboptimal solutions to the learning problem. For what regards the antisymmetry constraint, the present paper extends the setting of (Herbrich et al., 1998), by allowing for more general kernels, and providing several additional theoretical results (e.g., investigation of the relation between two different ways of imposing antisymmetry, construction of generic skew-balanced kernels starting from order-invariant kernels, antisymmetry preservation in suboptimal solutions to the binary classification problem, extension to regression and transdusctive SVMs). Finally, for what concerns the problem of arc classification on graphs, we recall that pairwise kernels are among the techniques that have been proposed for this purpose in the literature about social network analysis, other methods being based either on a Bayesian approach, or on a linear algebraic formulation (see the recent survey (Al Hasan and Zaki, 2011) for more details on all the three methods). In this context, an advantage of pairwise kernels is that they allow to integrate easily several features of possibly different nature, which can express, e.g., properties of the single nodes (individual features), or of the arcs joining them (group features). However, it has to be mentioned that all the pairwise kernel methods described in (Al Hasan and Zaki, 2011) refer only to the case of undirected arcs.

**Organization of the paper.** The paper is structured as follows. Section 2 illustrates briefly the kind of machine learning problems investigated in the paper. Section 3 lists some kernels that are used in the other sections. Section 4 investigates two methods to impose symmetry and antisymmetry constraints on the optimal solution of a kernel-based machine learning problem, focusing on the case of the training of an $l_1$-soft margin binary SVM classifier. Section 5 investigates a specific algorithm (one version of the Sequential Minimal Optimization (SMO) algorithm) from the point of view of symmetry/antisymmetry preservation at each of its iterations. Section 6 provides a numerical validation of the theoretical results. Section 7 provides several extensions of the analysis, namely, to support vector regression, to transductive SVMs, and to several graph kernels, including diffusion kernels. Finally, Section 8 concludes the paper. All the proofs are reported in the Appendix.

## 2. Symmetry and antisymmetry constraints

In this section, we provide a definition of the symmetry and antisymmetry constraints examined in the paper, together with an overview of the kind of problems examined.

In the following, given an ordered pair $(a, b)$ of objects $a, b$, belonging to some space of objects $\mathcal{O}$, and with $a \neq b$, we use the symbols $\underline{x}_a, \underline{x}_b \in \mathbb{R}^{n_1}$ to denote the row vectors of their individual features, and the symbol $\underline{x}_{ab} \in \mathbb{R}^{n_2}$ ($\underline{\tilde{x}}_{ab} \in \mathbb{R}^{n_3}$, respectively) to denote a row vector of group features that does not change at all (changes only in sign, respectively) after an exchange of order between the two objects of the ordered pair. Particularly simple examples of the three kinds of features are, respectively, position vectors, their Euclidean distance, and the relative position of the second object of the ordered pair with respect to the first one. We refer to (Dardard et al., 2016), for some more sophisticated examples of such individual and group features, i.e., those arising in the particular context of motion capture (see also footnote 23, reported in the last section).

With the notation above in mind, for $n = 2n_1 + n_2 + n_3$, we associate the feature vector

$$\underline{X}_{ab} := (\underline{x}_a, \underline{x}_{ab}, \underline{\tilde{x}}_{ab}, \underline{x}_b) \in \mathbb{R}^n \tag{1}$$

to the ordered pair $(a, b)$, and the feature vector

$$\underline{X}_{ba} := (\underline{x}_b, \underline{x}_{ba}, \underline{\tilde{x}}_{ba}, \underline{x}_a) = \mathcal{T}\underline{X}_{ab} \in \mathbb{R}^n \tag{2}$$

to the ordered pair $(b, a)$, where $\mathcal{T} : \mathbb{R}^n \to \mathbb{R}^n$ is the operator defined by

$$\mathcal{T}(\underline{x}_a, \underline{x}_{ab}, \underline{\tilde{x}}_{ab}, \underline{x}_b) := (\underline{x}_b, \underline{x}_{ab}, -\underline{\tilde{x}}_{ab}, \underline{x}_a).$$

Focusing for the moment on the case of binary classification, and denoting by $(c, d)$ another generic ordered pair of distinct objects $c$ and $d$, and by $f^\circ(\underline{X}_{cd})$ its classification produced by the trained classifier[1], one may want to impose, depending on the specific classification task, either the condition

$$f^\circ(\underline{X}_{cd}) = f^\circ(\underline{X}_{dc}) \quad (= f^\circ(\mathcal{T}\underline{X}_{cd})) \tag{3}$$

(symmetry constraint) or the condition

$$f^\circ(\underline{X}_{cd}) = -f^\circ(\underline{X}_{dc}) \quad (= -f^\circ(\mathcal{T}\underline{X}_{cd})) \tag{4}$$

(antisymmetry constraint). In case of a regression problem, $y_{ab}, y_{ba} \in \mathbb{R}$, $f^\circ$ denotes the obtained regression function, and the symmetry and antisymmetry constraints take the same form as above. It is worth remarking that, expressing the conditions (3) and (4) in terms of the operator $\mathcal{T}$, is closely related to the invariance framework of (Király et al., 2014). As an example, and as already mentioned in Section 1, the two constraints can be used, respectively, in pairwise classification, and in supervised learning of preference relations.

In the paper, we investigate how to impose these kinds of constraints (either at an optimal solution of the learning problem, or even at any element of a sequence of suboptimal solutions generated by a suitable algorithm), focusing on the case in which the function $f^\circ$ is obtained through a kernel method[2] (specifically, SVM classification and regression). As an example, in Section 4, taking the hint from (Brunner et al., 2012) (where the only condition (3) was considered), we consider two ways to impose condition (4) to the trained binary SVM classifier: the use of an antisymmetric training set (together with the choice of a suitable "order-invariant" kernel, see Section 4.1), and the use of a reduced training set (together with the choice of another suitable "skew-balanced" kernel, see Section 4.2). For completeness, in that section we report also analogous results (stated in terms of order-invariant and balanced kernels) related to the symmetry condition (3), which, for the particular problem considered therein, are specializations of the results obtained in (Brunner et al., 2012) to the case in which the operator $\mathcal{T}$ is used to relate the ordered pairs $(a, b)$ and $(b, a)$. In Section 5, we study how to impose either condition (3) or (4) to any element of the sequence of suboptimal solutions to the binary SVM classification problem generated by a particular optimization algorithm. Numerical results supporting the theory are provided in Section 6. Such results demonstrate the practical importance of imposing either condition (3) or (4) when the model generating the data labels satisfies it, and kernel methods are used. Finally, Section 7 reports extensions of the analysis to other kernel methods: in particular, Section 7.1 deals with the extension of the results contained in Section 4 to SVM regression, Section 7.2 provides an extension of the methodology introduced in Section 5 to the analysis of another optimization algorithm used to train transductive SVMs, whereas Sections 7.3 and Section 7.4 shows how to construct two different kinds of graph kernels inheriting the properties of the kernels considered in Section 4, as such properties are useful to impose either condition (3) or (4).

---

1. As it is usual in machine learning problems, we distinguish between the labels that are assigned externally by a teacher (denoted, in Section 4, by $y_{ab}, y_{ba} \in \{-1, +1\}$, for each of the two ordered pairs $(a, b)$ and $(b, a)$, respectively), and the classifications $f^\circ(\underline{X}_{cd}), f^\circ(\underline{X}_{dc}) \in \{-1, +1\}$ produced by the learning machine at the end of its training.

2. We recall that kernel methods are based on a function, called kernel, which makes it possible to express inner products between images of input vectors in a possibly infinite-dimensional feature space, obtained by applying a suitable (typically nonlinear) mapping. Then, a linear method (e.g., a linear classifier) is applied in that feature space. As such, kernel methods are able to generalize to the nonlinear case several other linear methods used in machine learning. Moreover, they are based on a strong theoretical foundation provided by Statistical Learning Theory (Vapnik, 1998), which allows one to derive bounds on the generalization error of the trained model.

**Remark 1** We conclude the section with the following considerations about individual and group features. In some cases, group features can be derived straightforwardly from individual features: e.g., if $\underline{x}_a$ and $\underline{x}_b$ are position vectors, then their Euclidean distance $x_{ab}$ has the expression $\|\underline{x}_b - \underline{x}_a\|_2$, where $\|\cdot\|_2$ is the $l_2$-norm, while the relative position of the second object with respect to the first one is $\underline{\tilde{x}}_{ab} = \underline{x}_b - \underline{x}_a$. Similarly, if a bag-of-features (also called bag-of-words (Shawe-Taylor and Cristianini, 2004)) approach is used (i.e., each individual feature represents the frequency of occurrence of some "word"), one could define simple group features such as sums and differences of corresponding individual features. More generally, by applying suitable kernels (e.g., the tensor pairwise learning kernel reported later in Section 3), one can map individual feature vectors to group feature vectors in the feature spaces $E$ associated with such kernels, which makes some examples of kernels reported in (Brunner et al., 2012) useful to deal with group feature vectors constructed starting from individual feature vectors. Finally, individual features can be interpreted as group features depending only on one object of the ordered pair. Then, one may conclude that only one kind of features (either individual or group features) is really needed. However, it is still useful to make a distinction between individual and group features, because group features could be derived from individual features also in a not straightforward way (this is, e.g., the case of the motion-related group features considered in (Dardard et al., 2016), which were obtained through several time-series analysis). In such a situation, it would be more natural to include them directly as inputs to the kernel, rather than to construct an equivalent kernel, which takes as inputs only the individual features, and deals with such specific group features implicitly in its feature space $E$, or using the mapping from the individual to the group features (if such a mapping is available), and the kernel itself results from the composition of such a mapping and a base kernel. Another case in which it is not possible to reduce group features to individual features is when the particular individual features from which such group features depend are not available (this does not exclude the case that, at the same time, other individual features, potentially useful for the specific machine learning problem, are still available).

## 3. Preliminaries: some typologies of pairwise kernels

To impose the symmetry and antisymmetry constraints (3) and (4), we need to introduce some basic notation about the following kinds of kernels, which are used extensively in the next sections:

(a) *pairwise kernel*: given a (possibly infinite-dimensional) Euclidean space $E$ (feature space) and a nonlinear mapping $\underline{\phi} : \mathbb{R}^n \to E$, the pairwise kernel $K : \mathbb{R}^n \times \mathbb{R}^n \to \mathbb{R}$ represents the inner product in the feature space $E$ between images of vectors in $\mathbb{R}^n$ under the mapping $\underline{\phi}$, according to the following definition:

$$K\left(\underline{X}_{ab}, \underline{X}_{cd}\right) := \left\langle \underline{\phi}\left(\underline{X}_{ab}\right), \underline{\phi}\left(\underline{X}_{cd}\right) \right\rangle_E.$$

It is also symmetric, in the sense that

$$K\left(\underline{X}_{ab}, \underline{X}_{cd}\right) = K\left(\underline{X}_{cd}, \underline{X}_{ab}\right),$$

which follows from the definition of inner product. Moreover, since it represents an inner product, it is positive semi-definite.

(b) *balanced kernel*: it is a pairwise kernel $K^b$ that satisfies the additional property

$$K^b\left(\underline{X}_{ab}, \underline{X}_{cd}\right) = K^b\left(\underline{X}_{ab}, \underline{X}_{dc}\right) \quad \left(= K^b\left(\underline{X}_{ab}, \mathcal{T}\underline{X}_{cd}\right)\right),$$

i.e., it is invariant under any exchange of order $(c, d) \to (d, c)$;

(c) *skew-balanced kernel*: it is a pairwise kernel $K^s$ that satisfies the additional property

$$K^s\left(\underline{X}_{ab}, \underline{X}_{cd}\right) = -K^s\left(\underline{X}_{ab}, \underline{X}_{dc}\right) \quad \left(= -K^s\left(\underline{X}_{ab}, \mathcal{T}\underline{X}_{cd}\right)\right),$$

i.e., it changes in sign under any exchange of order $(c, d) \to (d, c)$;

(d) *order-invariant kernel*: it is a pairwise kernel $K^o$ that satisfies the additional property

$$K^o\left(\underline{X}_{ab}, \underline{X}_{cd}\right) = K^o\left(\underline{X}_{ba}, \underline{X}_{dc}\right) \quad \left(= K^o\left(\mathcal{T}\underline{X}_{ab}, \mathcal{T}\underline{X}_{cd}\right)\right),$$

i.e., it is invariant under any two simultaneous exchanges of order $(a, b) \to (b, a)$ and $(c, d) \to (d, c)$.

Case (a) is just the application of the standard machine-learning definition of (symmetric and positive semi-definite) kernel (Shawe-Taylor and Cristianini, 2004) to the situation in which each feature vector represents an ordered pair of objects (rather than, e.g., a single object), whereas cases (b), (c), and (d) impose additional invariance properties (cases (b), and (d) also follow as special cases of the invariance framework considered in (Király et al., 2014)). Moreover, cases (a), (b) and (d) are specifications (due to the inclusion of both individual and group features, and the use of the operator $\mathcal{T}$) of the corresponding cases investigated in (Brunner et al., 2012)[3]., whereas case (c) extends in a similar way the one recently introduced independently in (Pahikkala et al., 2015)[4].

As already observed in (Brunner et al., 2012), it follows from the definitions of pairwise kernels, balanced kernels, and order-invariant kernels, that every balanced kernel is also an order-invariant kernel. Some examples of pairwise kernels that are order-invariant but not balanced (according to the definitions reported above in this section, in which $\underline{X}_{ab}$, $\underline{X}_{ba}$ and $\underline{X}_{cd}$, $\underline{X}_{dc}$ are further related by $\underline{X}_{ba} = \mathcal{T}\underline{X}_{ab}$ and $\underline{X}_{dc} = \mathcal{T}\underline{X}_{cd}$) are the linear kernel:

$$
\begin{aligned}
& K^o_{\text{lin}}\left(\underline{X}_{ab}, \underline{X}_{cd}\right) \\
:= \quad & \langle \underline{x}_a, \underline{x}_c \rangle_{\mathbb{R}^{n_1}} + \langle \underline{x}_{ab}, \underline{x}_{cd} \rangle_{\mathbb{R}^{n_2}} + \langle \underline{\tilde{x}}_{ab}, \underline{\tilde{x}}_{cd} \rangle_{\mathbb{R}^{n_3}} + \langle \underline{x}_b, \underline{x}_d \rangle_{\mathbb{R}^{n_1}} \\
= \quad & \langle \underline{x}_b, \underline{x}_d \rangle_{\mathbb{R}^{n_1}} + \langle \underline{x}_{ab}, \underline{x}_{cd} \rangle_{\mathbb{R}^{n_2}} + \langle (-\underline{\tilde{x}}_{ab}), (-\underline{\tilde{x}}_{cd}) \rangle_{\mathbb{R}^{n_3}} + \langle \underline{x}_a, \underline{x}_c \rangle_{\mathbb{R}^{n_1}} \\
= \quad & \langle \underline{x}_b, \underline{x}_d \rangle_{\mathbb{R}^{n_1}} + \langle \underline{x}_{ba}, \underline{x}_{dc} \rangle_{\mathbb{R}^{n_2}} + \langle \underline{\tilde{x}}_{ba}, \underline{\tilde{x}}_{dc} \rangle_{\mathbb{R}^{n_3}} + \langle \underline{x}_a, \underline{x}_c \rangle_{\mathbb{R}^{n_1}} \\
= \quad & K^o_{\text{lin}}\left(\underline{X}_{ba}, \underline{X}_{dc}\right),
\end{aligned}
$$

the homogeneous polynomial kernel of order $m$ (where $m$ is a positive integer):

$$
K^o_{\text{pol}_d}\left(\underline{X}_{ab}, \underline{X}_{cd}\right) := \left(K^o_{\text{lin}}\left(\underline{X}_{ab}, \underline{X}_{cd}\right)\right)^m, \tag{5}
$$

and, for $\sigma > 0$, the Gaussian kernel:

$$
\begin{aligned}
& K^o_{\text{Gauss}}\left(\underline{X}_{ab}, \underline{X}_{cd}\right) \\
:= \quad & \exp\left( -\frac{\left( \|\underline{x}_a - \underline{x}_c\|_2^2 + \|\underline{x}_{ab} - \underline{x}_{cd}\|_2^2 + \|\underline{\tilde{x}}_{ab} - \underline{\tilde{x}}_{cd}\|_2^2 + \|\underline{x}_b - \underline{x}_d\|_2^2 \right)}{2\sigma^2} \right).
\end{aligned}
$$

---

3. We have used the term order-invariant for the case (d), for which no specific term was used in (Brunner et al., 2012). The term permutation invariant is used in (Pahikkala et al., 2015).

4. In (Pahikkala et al., 2015), this kernel has been introduced for a different investigation than ours, as it is related to spectral properties of a suitable linear operator associated with the kernel.

It is worth remarking that linear, polynomial and Gaussian kernels are also considered both in (Brunner et al., 2012) and in (Király et al., 2014). An example of a pairwise kernel that is not order-invariant is given later in Section 6.

Likewise in (Brunner et al., 2012), starting from any order-invariant kernel $K^o$ (associated with the mapping $\underline{\phi}^o$) and exploiting its symmetry, it is possible to define a balanced kernel $K^b$ in the following way:

$$
\begin{aligned}
& K^b\left(\underline{X}_{ab}, \underline{X}_{cd}\right) \\
:= \quad & \frac{1}{2}\left(K^o\left(\underline{X}_{ab}, \underline{X}_{cd}\right) + K^o\left(\underline{X}_{ba}, \underline{X}_{cd}\right)\right) \\
= \quad & \frac{1}{4}\left(K^o\left(\underline{X}_{ab}, \underline{X}_{cd}\right) + K^o\left(\underline{X}_{ba}, \underline{X}_{cd}\right) + K^o\left(\underline{X}_{ba}, \underline{X}_{dc}\right) + K^o\left(\underline{X}_{ab}, \underline{X}_{dc}\right)\right) \\
= \quad & \left\langle \frac{1}{2}\left(\underline{\phi}^o\left(\underline{X}_{ab}\right) + \underline{\phi}^o\left(\underline{X}_{ba}\right)\right), \frac{1}{2}\left(\underline{\phi}^o\left(\underline{X}_{cd}\right) + \underline{\phi}^o\left(\underline{X}_{dc}\right)\right) \right\rangle_E,
\end{aligned} \tag{6}
$$

where the first equality above is obtained by exploiting the order invariance of $K^o$, and the second one by using the definition of pairwise kernel, and collecting terms. The last line in (6) shows that $K^b$ is associated with the mapping $\underline{\phi}^b : \mathbb{R}^n \to E$ defined by

$$
\underline{\phi}^b\left(\underline{X}_{ab}\right) := \frac{1}{2}\left(\underline{\phi}^o\left(\underline{X}_{ab}\right) + \underline{\phi}^o\left(\underline{X}_{ba}\right)\right),
$$

hence, $K^b$ really expresses an inner product in the feature space $E$, and is symmetric.

Similarly, starting from the order-invariant kernel $K^o$, it is possible to define a skew-balanced kernel $K^s$ in the following way:

$$
\begin{aligned}
& K^s\left(\underline{X}_{ab}, \underline{X}_{cd}\right) \\
:={}& \frac{1}{2}\left(K^o\left(\underline{X}_{ab}, \underline{X}_{cd}\right) - K^o\left(\underline{X}_{ba}, \underline{X}_{cd}\right)\right) \\
={}& \frac{1}{4}\left(K^o\left(\underline{X}_{ab}, \underline{X}_{cd}\right) - K^o\left(\underline{X}_{ba}, \underline{X}_{cd}\right) - K^o\left(\underline{X}_{ab}, \underline{X}_{dc}\right) + K^o\left(\underline{X}_{ba}, \underline{X}_{dc}\right)\right) \\
={}& \left\langle \frac{1}{2}\left(\underline{\phi}^o\left(\underline{X}_{ab}\right) - \underline{\phi}^o\left(\underline{X}_{ba}\right)\right), \frac{1}{2}\left(\underline{\phi}^o\left(\underline{X}_{cd}\right) - \underline{\phi}^o\left(\underline{X}_{dc}\right)\right)\right\rangle_E,
\end{aligned}
\tag{7}
$$

where the last line shows that $K^s$ is associated with the mapping $\underline{\phi}^s : \mathbb{R}^n \to E$ defined by

$$
\underline{\phi}^s\left(\underline{X}_{ab}\right) := \frac{1}{2}\left(\underline{\phi}^o\left(\underline{X}_{ab}\right) - \underline{\phi}^o\left(\underline{X}_{ba}\right)\right).
$$

Hence, $K^s$ really expresses an inner product in the feature space $E$, and is symmetric. To exclude the trivial kernel $K^s \equiv 0$, the order-invariant kernel $K^o$ in (7) is required not to be balanced (which is, as already observed, a particular case of order-invariant kernel).

In the following sections, to simplify the notation, in a similar way as in (Brunner et al., 2012), we use the shortcuts $K^o_{ab,cd} := K^o\left(\underline{X}_{ab}, \underline{X}_{cd}\right)$, $K^b_{ab,cd} := K^b\left(\underline{X}_{ab}, \underline{X}_{cd}\right)$, and $K^s_{ab,cd} := K^s\left(\underline{X}_{ab}, \underline{X}_{cd}\right)$.

**Remark 2** Other pairwise kernels that are also balanced (or skew-balanced) according to the definitions reported in the paper can be constructed by composition with standard kernels, applied to subvectors of the feature vectors. More precisely, given a symmetric positive semi-definite kernel $k_1 : \mathbb{R}^{n_1} \times \mathbb{R}^{n_1} \to \mathbb{R}$ applied to the individual feature vectors $\underline{x}_a$, $\underline{x}_b$, $\underline{x}_c$, $\underline{x}_d$, starting from the balanced kernel

$$
K^b_{DL}\left(\left(\underline{x}_a, \underline{x}_b\right), \left(\underline{x}_c, \underline{x}_d\right)\right) := \frac{1}{2}\left(k_1(\underline{x}_a, \underline{x}_c) + k_1(\underline{x}_a, \underline{x}_d) + k_1(\underline{x}_b, \underline{x}_c) + k_1(\underline{x}_b, \underline{x}_d)\right)
$$

(called *direct sum learning pairwise kernel* (Bar-Hillel et al., 2004)), one can define the balanced kernel

$$
\hat{K}^b_{DL}\left(\underline{X}_{ab}, \underline{X}_{cd}\right) := K^b_{DL}\left(\left(\underline{x}_a, \underline{x}_b\right), \left(\underline{x}_c, \underline{x}_d\right)\right) + k_2\left(\underline{x}_{ab}, \underline{x}_{cd}\right) + k_3\left(\underline{\tilde{x}}_{ab}, \underline{\tilde{x}}_{cd}\right),
\tag{8}
$$

where $k_2 : \mathbb{R}^{n_2} \times \mathbb{R}^{n_2} \to \mathbb{R}$ and $k_3 : \mathbb{R}^{n_3} \times \mathbb{R}^{n_3} \to \mathbb{R}$ are symmetric positive semi-definite kernels, with $k_3$ satisfying the additional property

$$
k_3\left(\underline{\tilde{x}}_{ab}, \underline{\tilde{x}}_{cd}\right) = k_3\left(\underline{\tilde{x}}_{ab}, \underline{\tilde{x}}_{dc}\right) = k_3\left(\underline{\tilde{x}}_{ab}, -\underline{\tilde{x}}_{cd}\right)
\tag{9}
$$

(which holds, e.g., when $k_3$ is a homogeneous polynomial kernel of even order $m$, see formula (5)). Similarly, one can also define the skew-balanced kernel

$$
\hat{K}^s_{DL}\left(\underline{X}_{ab}, \underline{X}_{cd}\right) := \left(K^b_{DL}\left(\left(\underline{x}_a, \underline{x}_b\right), \left(\underline{x}_c, \underline{x}_d\right)\right) + k_2\left(\underline{x}_{ab}, \underline{x}_{cd}\right)\right) k_3\left(\underline{\tilde{x}}_{ab}, \underline{\tilde{x}}_{cd}\right),
\tag{10}
$$

if $k_3$ satisfies, instead of (9), the additional property

$$
k_3\left(\underline{\tilde{x}}_{ab}, \underline{\tilde{x}}_{cd}\right) = -k_3\left(\underline{\tilde{x}}_{ab}, \underline{\tilde{x}}_{dc}\right) = -k_3\left(\underline{\tilde{x}}_{ab}, -\underline{\tilde{x}}_{cd}\right)
\tag{11}
$$

(which holds, e.g., when $k_3$ is a homogeneous polynomial kernel of odd order $m$, see again formula (5)). Finally, similar constructions of balanced/skew-balanced kernels as (8) and (10) can be made, starting from the balanced kernels

$$K_{TL}\left((\underline{x}_a, \underline{x}_b), (\underline{x}_c, \underline{x}_d)\right) := \frac{1}{2}\left(k_1(\underline{x}_a, \underline{x}_c)k_1(\underline{x}_b, \underline{x}_d) + k_1(\underline{x}_a, \underline{x}_d)k_1(\underline{x}_b, \underline{x}_c)\right),$$

$$K_{ML}\left((\underline{x}_a, \underline{x}_b), (\underline{x}_c, \underline{x}_d)\right) := \frac{1}{4}\left(k_1(\underline{x}_a, \underline{x}_c) - k_1(\underline{x}_a, \underline{x}_d) - k_1(\underline{x}_b, \underline{x}_c) + k_1(\underline{x}_b, \underline{x}_d)\right)^2,$$

$$K_{TM}\left((\underline{x}_a, \underline{x}_b), (\underline{x}_c, \underline{x}_d)\right) := K_{TL}\left((\underline{x}_a, \underline{x}_b), (\underline{x}_c, \underline{x}_d)\right) + K_{ML}\left((\underline{x}_a, \underline{x}_b), (\underline{x}_c, \underline{x}_d)\right)$$

(called, respectively, *tensor learning pairwise kernel* (Vert et al., 2007), *metric learning pairwise kernel* (Vert et al., 2007), and *tensor metric learning pairwise kernel* (Brunner et al., 2012)).

## 4. Two methods to impose symmetry and antisymmetry constraints on the optimal solution of a kernel-based machine learning problem

In the following, we illustrate two methods to impose symmetry and antisymmetry constraints on the optimal solution of a kernel-based machine learning problem. Specifically, we focus on the training of an $l_1$-soft margin binary SVM classifier(Shawe-Taylor and Cristianini, 2004, Section 7.2.2) (see Section 7 for extensions).

### 4.1 Use of symmetric/antisymmetric training sets

We denote by $I$ the set of ordered pairs $(a, b)$ used in the training of the learning machine. The label associated with the ordered pair $(a, b)$ is denoted by $y_{ab} \in \{-1, 1\}$. In the following, we assume that every time the training set $I$ contains the ordered pair $(a, b)$, it also contains the ordered pair $(b, a)$. Moreover, we assume that $I$ does not contain ordered pairs of the form $(a, a)$. For $C > 0$ and an order-invariant kernel $K^o$ associated with the mapping $\underline{\phi}^o$ in the feature space $E$, the primal optimization problem that models the training of the corresponding $l_1$-soft margin binary SVM classifier is:

$$\text{minimize}_{\underline{w} \in E, \gamma \in \mathbb{R}, \{\xi_{ab} \in \mathbb{R} : (a,b) \in I\}} \quad \frac{1}{2}\|\underline{w}\|_E^2 + C \sum_{(a,b) \in I} \xi_{ab},$$

$$\text{s.t.} \quad y_{ab}\left(\left\langle \underline{w}, \underline{\phi}^o\left(\underline{X}_{ab}\right)\right\rangle_E + \gamma\right) \geq 1 - \xi_{ab}, \quad \forall (a, b) \in I,$$

$$\xi_{ab} \geq 0, \quad \forall (a, b) \in I \tag{12}$$

(see also (Shawe-Taylor and Cristianini, 2004, Section 7.2.2)). The corresponding dual optimization problem is

$$\text{minimize}_{\{\alpha_{ab} \in \mathbb{R} : (a,b) \in I\}} \quad G(\underline{\alpha}),$$

$$\text{s.t.} \quad 0 \leq \alpha_{ab} \leq C, \quad \forall (a, b) \in I,$$

$$\sum_{(a,b) \in I} y_{ab}\alpha_{ab} = 0, \tag{13}$$

where

$$G(\underline{\alpha}) := \frac{1}{2} \sum_{(a,b),(c,d) \in I} \alpha_{ab}\alpha_{cd}y_{ab}y_{cd}K^o_{ab,cd} - \sum_{(a,b) \in I} \alpha_{ab}. \tag{14}$$

Both problems (12) and (13) are convex quadratic optimization problems[5]. By Weierstrass theorem, each of them admits an optimal solution. It is well-known from the theory of SVMs that any optimal weight vector $\underline{w}^\circ$ of the primal optimization problem (12) is expressed in terms of support vectors, i.e., vectors $\underline{\phi}^o\left(\underline{x}_{ab}\right)$ associated with optimal dual variables $\alpha_{ab}^\circ$ that are different from 0 at dual optimality. More precisely, starting from any optimal solution $\underline{\alpha}^\circ$ to the dual optimization problem (13), one can construct an optimal weight vector $\underline{w}^\circ$ for the primal optimization problem (12) as follows[6]:

$$\underline{w}^\circ := \sum_{(a,b)\in I} \alpha_{ab}^\circ y_{ab} \underline{\phi}^o\left(\underline{x}_{ab}\right).\tag{15}$$

Additionally, it follows from (Burges and Crisp, 2000, Theorem 2) that, even in case of nonuniqueness of the optimal solution to the primal optimization problem (12), the optimal weight vector $\underline{w}^\circ$ is any case unique (nonuniqueness of the optimal solution to such a problem may arise from the only fact that the set of optimal values for the bias $\gamma$ is a closed interval of the real line). The representation (15) is well-known as the representer theorem for $l_1$-soft margin binary SVM classifiers (see, e.g., (Shawe-Taylor and Cristianini, 2004, Section 7.2.2)).

The following lemma, which is used in the proof of the next Theorem 1, specializes (Brunner et al., 2012, Lemma 1) to the case of the specific order-invariant kernels considered in this paper, for which $\underline{X}_{ab}$, $\underline{X}_{ba}$ and $\underline{X}_{cd}, \underline{X}_{dc}$ are further related by $\underline{X}_{ba} = \mathcal{T}\underline{X}_{ab}$ and $\underline{X}_{dc} = \mathcal{T}\underline{X}_{cd}$. Moreover, differently from (Brunner et al., 2012, Lemma 1), the result also considers the case of labels satisfying, for every $(a,b) \in I$, the antisymmetry condition $y_{ab} = -y_{ba}$. The lemma provides a specific instance of the representer theorem for training an $l_1$-soft margin binary SVM classifier with a symmetric/antisymmetric training set and an order-invariant kernel, by constraining the coefficients in formula (15) to satisfy $\alpha_{ab}^\circ = \alpha_{ba}^\circ$, for all the elements of the training set.

**Lemma 1** *Let $K^o$ be an order-invariant kernel and one of the following conditions hold:*

*(a) for all $(a,b) \in I$, $y_{ab} = y_{ba}$ (symmetry condition);*

*(b) for all $(a,b) \in I$, $y_{ab} = -y_{ba}$ (antisymmetry condition).*

*Then, there exists an optimal solution $\underline{\alpha}^\circ$ to the dual optimization problem (13) for which*

$$\alpha_{ab}^\circ = \alpha_{ba}^\circ, \quad \forall (a,b) \in I.\tag{16}$$

Given an optimal solution $(\underline{w}^\circ, \gamma^\circ, \{\xi_{ab}^\circ \in \mathbb{R} : (a,b) \in I\})$ to the primal optimization problem (12), let

$$f^\circ(\underline{X}_{cd}) := \mathrm{sgn}\left(\left\langle \underline{w}^\circ, \underline{\phi}^o\left(\underline{X}_{cd}\right)\right\rangle_E + \gamma^\circ\right)\tag{17}$$

be the associated classification function, where sgn is the signum function, defined[7] as

$$\mathrm{sgn}(z) := \begin{cases} -1, & \text{if } z < 0, \\ 0, & \text{if } z = 0, \\ 1, & \text{if } z > 0. \end{cases}$$

---

5. One difference in our formulation of the dual optimization problem with respect to (Brunner et al., 2012) is the absence of the constraint $0 \le \alpha_{aa} \le 2C$, since the set $I$ does not contain elements of the form $(a,a)$.

6. The result follows by looking for a stationary point of the Lagrangian associated with the primal optimization problem (12), and particularly, by setting to $\underline{0}$ its gradient vector with respect to the weight vector $\underline{w}$ (see, e.g., (Shawe-Taylor and Cristianini, 2004)).

7. For symmetry reasons, we define $\mathrm{sgn}(0) := 0$ (so, formally, the classification function (17) is not binary in this case, but one can interpret as degenerate the case in which the argument of sgn is 0).

The following result extends (Brunner et al., 2012, Theorem 2) to the present case. It provides sufficient conditions under which an optimal solution to the primal optimization problem (12) satisfies the symmetry (respectively, antisymmetry) constraint. Again, for the symmetric case, the result is just a specialization of (Brunner et al., 2012, Theorem 2) to the specific ordered-invariant kernels considered in the paper, whereas the antisymmetric case is new.

**Theorem 1** *The following hold.*

(a) *Let $K^o$ be an order-invariant kernel and, for all $(a, b) \in I$, let $y_{ab} = y_{ba}$. Then, for any ordered pair $(c, d)$ (even outside the training set), all the optimal solutions to the primal optimization problem (12) satisfy*

$$f^\circ(\underline{X}_{cd}) = f^\circ(\underline{X}_{dc}) \ \ (= f^\circ(\mathcal{T}\underline{X}_{cd})). \tag{18}$$

(b) *Let $K^o$ be an order-invariant kernel and, for all $(a, b) \in I$, let $y_{ab} = -y_{ba}$. Then, there exists an optimal solution to the primal optimization problem (12) such that, for any ordered pair $(c, d)$ (even outside the training set), one has*

$$f^\circ(\underline{X}_{cd}) = -f^\circ(\underline{X}_{dc}) \ \ (= -f^\circ(\mathcal{T}\underline{X}_{cd})). \tag{19}$$

### 4.2 Use of balanced/skew-balanced kernels

An alternative way to impose the symmetry/antisymmetry conditions of Theorem 1 to an optimal solution of the primal optimization problem (12) consists in solving, respectively, the following two variations (a) and (b) of the dual optimization problem (13). In the following, for every unordered pair of distinct objects $a$, $b$, we fix one of the two possible orders (say, either $(a, b)$ or $(b, a)$, but not both), and we denote by $J$ the set of ordered pairs obtained by this procedure.

Let $K^o$ be any order-invariant kernel, and consider the following optimization problems.

(a) Case of symmetric labels:

$$\begin{aligned}
\text{minimize}_{\{\beta_{ab} \in \mathbb{R} : (a,b) \in J\}} \quad & H^b(\underline{\beta}), \\
\text{s.t.} \quad & 0 \leq \beta_{ab} \leq 2C, \quad \forall (a, b) \in J, \\
& \sum_{(a,b) \in J} y_{ab}\beta_{ab} = 0,
\end{aligned} \tag{20}$$

where $H^b(\underline{\beta}) := \frac{1}{2} \sum_{(a,b),(c,d) \in J} \beta_{ab}\beta_{cd}y_{ab}y_{cd}K^b_{ab,cd} - \sum_{(a,b) \in J} \beta_{ab}$, and

$$K^b_{ab,cd} := \frac{1}{2} \left( K^o_{ab,cd} + K^o_{ba,cd} \right). \tag{21}$$

(b) Case of antisymmetric labels:

$$\begin{aligned}
\text{minimize}_{\{\beta_{ab} \in \mathbb{R} : (a,b) \in J\}} \quad & H^s(\underline{\beta}), \\
\text{s.t.} \quad & 0 \leq \beta_{ab} \leq 2C, \quad \forall (a, b) \in J,
\end{aligned} \tag{22}$$

where $H^s(\underline{\beta}) := \frac{1}{2} \sum_{(a,b),(c,d) \in J} \beta_{ab}\beta_{cd}y_{ab}y_{cd}K^s_{ab,cd} - \sum_{(a,b) \in J} \beta_{ab}$, and

$$K^s_{ab,cd} := \frac{1}{2} \left( K^o_{ab,cd} - K^o_{ba,cd} \right). \tag{23}$$

The two kernels $K^b$ and $K^s$ in (21) and (23) are, respectively, the balanced kernel defined in (6), and the skew-balanced kernel defined in (7). Both problems (20) and (22) are convex quadratic optimization problems. By Weierstrass theorem, each of them admits an optimal solution. It is also interesting to observe that the optimization problem (20) is the dual of a primal optimization problem of the form (12), with $I$ replaced by $J$, $C$ by $2C$, and $\underline{\phi}^o$ by $\underline{\phi}^b$, whereas the optimization problem (22) is the dual of a primal optimization problem of the form (12) but without the bias $\gamma$ as an optimization variable (more precisely, with such a bias set to 0)[8], and with $I$ replaced by $J$, $C$ by $2C$, and $\underline{\phi}^o$ by $\underline{\phi}^s$. For this reason, likewise in Section 4.1, the classical representer theorem of $l_1$-soft margin binary SVM classifiers applies without changes, i.e., the optimal weight vectors to both primal optimization problems are represented in terms of support vectors.

The following specializes (Brunner et al., 2012, Theorem 3) to the specific order-invariant kernels of this paper in the symmetric case, and extends in the antisymmetric case. It shows that, in order to obtain the optimal classifier (17) - apart from its optimal bias[9] $\gamma^\circ$ - for the case of order-invariant kernels and symmetric and antisymmetric labels, respectively, one can solve the dual optimization problem (13) or equivalently, respectively, the optimization problems (20) and (22).

**Theorem 2** *The following hold.*

(a) *Let $\underline{\alpha}^\circ$ and $\underline{\beta}^\circ$ denote optimal solutions, respectively, to the dual optimization problem (13) and to the optimization problem (20). Then, for any ordered pair $(c,d)$ of objects (even outside the training set), one has*

$$\left\langle \underline{w}^\circ, \underline{\phi}^o\left(\underline{X}_{cd}\right)\right\rangle_E = \sum_{(a,b)\in I} \alpha_{ab}^\circ y_{ab} K_{ab,cd}^o = \sum_{(a,b)\in J} \beta_{ab}^\circ y_{ab} K_{ab,cd}^b. \tag{24}$$

*Moreover, the optimal values of the objectives of the optimization problems (13) and (20) are the same.*

(b) *Let $\underline{\alpha}^\circ$ and $\underline{\beta}^\circ$ denote optimal solutions, respectively, to the dual optimization problem (13) and to the optimization problem (22). Then, for any ordered pair $(c,d)$ of objects (even outside the training set), one has*

$$\left\langle \underline{w}^\circ, \underline{\phi}^o\left(\underline{X}_{cd}\right)\right\rangle_E = \sum_{(a,b)\in I} \alpha_{ab}^\circ y_{ab} K_{ab,cd}^o = \sum_{(a,b)\in J} \beta_{ab}^\circ y_{ab} K_{ab,cd}^s. \tag{25}$$

*Moreover, the optimal values of the objectives of the optimization problems (13) and (22) are the same.*

**Remark 3** From a computational point of view, the optimization problems (20) and (22) involve one half of the (ordered pairs of) training examples needed to solve the dual optimization problem (13). Moreover, the kernel matrices of (20) and (22) have a number of elements which is four times smaller than the one of (13), even though the evaluation of $K^b$ and $K^s$ involves two evaluations of $K^o$. To save memory space, one can cache the individual features (likewise in (Brunner et al., 2012)), then compute the kernel values every time they are needed, only if the group features can be easily computed from the pairs of individual features. Otherwise, one has either to store also the group features, or to store the whole kernel matrix associated with the given set of (ordered pairs of) training examples. Since each ordered pair can be interpreted as a directed arc on a graph, in practice the last two approaches are feasible only for sparse graphs, i.e., graphs whose number of edges is, e.g., linear with respect to the number of nodes.

---

8. Indeed, the condition $\sum_{(a,b)\in I} y_{ab}\alpha_{ab} = 0$ in the dual optimization problem (13) arises from setting to 0 the partial derivative with respect to $\gamma$ of the Lagrangian associated with the primal optimization problem (12) (see, e.g., (Shawe-Taylor and Cristianini, 2004)). Such a condition is absent if $\gamma$ is not included as an optimization variable in the primal optimization problem (12).

9. However, once the term $\left\langle \underline{w}^\circ, \underline{\phi}^o\left(\underline{X}_{cd}\right)\right\rangle_E$ has been computed, it is well-known that one can easily find an optimal bias $\gamma^\circ$ in (12), e.g., by applying the Karush-Kuhn-Tucker (KKT) conditions.

## 5. Analysis of the Sequential Minimal Optimization algorithm from the point of view of symmetry/antisymmetry preservation at each iteration

In this section, we focus on the investigation of how it is possible to obtain either symmetric or antisymmetric suboptimal solutions to the primal and dual optimization problems (12) and (13), focusing on the behavior of a commonly-used algorithm to solve the dual problem (13), which is the Sequential Minimal Optimization (SMO) algorithm (Platt, 1999). We recall that, in each of its iterations, both primal and dual feasibility are guaranteed, whereas the Karush-Kuhn-Tucker (KKT) conditions may be violated in case of suboptimality of the current solution. At each iteration $t \geq 1$, such an algorithm fixes all the dual variables to the previous values except from two of them, then solves the resulting two-variable optimization problem (in closed form, which is one of the main advantages of the algorithm with respect to other ones), resulting in a new feasible dual solution, which improves the value of the objective when the previous feasible dual solution is suboptimal. A modified version of such an algorithm, based on a suitable selection of the two dual variables to be re-optimized at each iteration (Keerthi et al., 2001), is guaranteed to converge asymptotically to an optimal solution of the dual problem (13) (Lin, 2001, 2002) and has been implemented in the software LIBSVM (Chang and Lin, 2011). More precisely, in this case, adapting the notation to the dual problem (13), the dual variables $\alpha_{a^{(t,1)}b^{(t,1)}}, \alpha_{a^{(t,2)}b^{(t,2)}}$ selected at the iteration $t$ are the ones that have the maximal violation of the corresponding KKT condition, according to the following formula (see (Lin, 2002, formula (2))):

$$
\begin{aligned}
\left(a^{(t,1)}, b^{(t,1)}\right) &\in \operatorname{argmax}\left(\left\{-F\left(\alpha_{ab}^{(t-1)}\right) : y_{ab} = 1, \alpha_{ab}^{(t-1)} < C\right\}, \left\{F\left(\alpha_{ab}^{(t-1)}\right) : y_{ab} = -1, \alpha_{ab}^{(t-1)} > 0\right\}\right), \\
\left(a^{(t,2)}, b^{(t,2)}\right) &\in \operatorname{argmin}\left(\left\{F\left(\alpha_{ab}^{(t-1)}\right) : y_{ab} = -1, \alpha_{ab}^{(t-1)} < C\right\}, \left\{-F\left(\alpha_{ab}^{(t-1)}\right) : y_{ab} = 1, \alpha_{ab}^{(t-1)} > 0\right\}\right),
\end{aligned}
\tag{26}
$$

where

$$
F\left(\alpha_{ab}^{(t-1)}\right) := \sum_{(c,d) \in I} y_{ab} y_{cd} K_{ab,cd}^o \alpha_{cd}^{(t-1)} - 1,
\tag{27}
$$

and $\underline{\alpha}^{(t-1)}$ denotes the feasible dual solution coming from the previous iteration.

In this framework, it is interesting to examine whether, starting from any initial feasible dual solution $\underline{\alpha}^{(0)}$ satisfying the symmetry condition

$$
\alpha_{ab}^{(0)} = \alpha_{ba}^{(0)}, \quad \forall (a,b) \in I,
\tag{28}
$$

(which holds, e.g., if $\underline{\alpha}^{(0)} = \underline{0}$, which is always feasible for the dual problem (13)), a similar symmetry condition

$$
\alpha_{ab}^{(t)} = \alpha_{ba}^{(t)}, \quad \forall (a,b) \in I,
\tag{29}
$$

holds at each iteration $t$ of the version of the SMO algorithm described by the selection rule (26) above (possibly applied, in the symmetric case, not directly to the dual problem (13), but to the equivalent one (20), since it has a similar form). In such a case, also the primal feasible solution obtained at the iteration $t$, i.e.,

$$\underline{w}^{(t)} := \sum_{(a,b) \in I} \alpha_{ab}^{(t)} y_{ab} \underline{\phi}^o \left( \underline{X}_{ab} \right), \tag{30}$$

and the corresponding bias $\gamma^{(t)}$ determined at the same iteration (Platt, 1999, Section 12.2.3), are such that the associated classification function

$$f^{(t)}(\underline{X}_{cd}) := \text{sgn} \left( \left\langle \underline{w}^{(t)}, \underline{\phi}^o \left( \underline{X}_{cd} \right) \right\rangle_E + \gamma^{(t)} \right) \tag{31}$$

satisfies, for any ordered pair of objects $(c, d)$,

$$f^{(t)}(\underline{X}_{cd}) = f^{(t)}(\underline{X}_{dc}) \quad \left( = f^{(t)}(\mathcal{T}\underline{X}_{cd}) \right) \tag{32}$$

if, for all $(a, b) \in I$, one has $y_{ab} = y_{ba}$, or

$$f^{(t)}(\underline{X}_{cd}) = -f^{(t)}(\underline{X}_{dc}) \quad \left( = -f^{(t)}(\mathcal{T}\underline{X}_{cd}) \right) \tag{33}$$

if, for all $(a, b) \in I$, one has $y_{ab} = -y_{ba}$. As it is shown in our following result, one can indeed make the condition (29) hold for every iteration $t$.

**Theorem 3** *The following hold.*

(a) *Let $K^o$ be an order-invariant kernel and, for all $(a, b) \in I$, let $y_{ab} = y_{ba}$. Let the version of the SMO algorithm described by the selection rule (26) be applied to the optimization problem (20), and denote by $\underline{\beta}^{(t)}$ its solution generated at the iteration $t$. Moreover, let $\underline{\alpha}^{(t)}$ be defined as the vector with components*

$$\alpha_{ab}^{(t)} := \begin{cases} \frac{\beta_{ab}^{(t)}}{2}, & \text{if } (a, b) \in J, \\ \frac{\beta_{ba}^{(t)}}{2}, & \text{if } (b, a) \in J. \end{cases} \tag{34}$$

*Then, at each iteration $t$, $\underline{\alpha}^{(t)}$ is feasible for the dual optimization problem (13), and satisfies the symmetry condition (29). Moreover, when $t$ tends to $+\infty$, the sequence of the $\underline{\alpha}^{(t)}$'s converges to an optimal solution of the dual problem (13).*

(b) *Let $K^o$ be an order-invariant kernel and, for all $(a, b) \in I$, let $y_{ab} = -y_{ba}$. Let the version of the SMO algorithm described by the selection rule (26) be applied to the optimization problem (13), and suppose that $\underline{\alpha}^{(0)}$ satisfies the symmetry condition (28). Then, at each iteration $t$, it is possible to choose $\left( a^{(t,1)}, b^{(t,1)} \right)$ and $\left( a^{(t,2)}, b^{(t,2)} \right)$ according to formula (26) in such a way that $\underline{\alpha}^{(t)}$ satisfies the symmetry condition (29). Moreover, when $t$ tends to $+\infty$, the sequence of the $\underline{\alpha}^{(t)}$'s converges to an optimal solution of the dual problem (13).*

**Remark 4** One can notice that Theorem 3 (b) does not preclude the existence of other choices of $\left( a^{(t,1)}, b^{(t,1)} \right)$ and $\left( a^{(t,2)}, b^{(t,2)} \right)$, still selected according to formula (26), for which $\underline{\alpha}^{(t)}$ does not satisfy the symmetry condition (29). In particular, this could happen when the argmax and argmin sets in formula (26) contain more than one element.

**Remark 5** Theorem 3 shows conditions under which the sequence of suboptimal solutions obtained applying the SMO algorithm satisfies symmetry/antisymmetry properties valid also for the optimal solution. Associated rates of convergence could be obtained by exploiting related results valid for the SMO algorithm. For instance, (Chen et al., 2006) proved linear convergence of SMO under certain assumptions.

**Remark 6** The following is an alternative way to impose the symmetry condition (29) after every iteration of the SMO algorithm with the selection rule (26), when it is applied to the dual optimization problem (13). After generating the possibly asymmetric feasible dual solution $\underline{\alpha}^{(t)}$, one constructs another feasible dual solution $\underline{\bar{\alpha}}^{(t)}$ defined by $\bar{\alpha}_{ab}^{(t)} := \alpha_{ba}^{(t)}$, for all $(a, b) \in I$, then generates the feasible dual solution $\underline{\bar{\bar{\alpha}}}^{(t)}$ defined by $\underline{\bar{\bar{\alpha}}}^{(t)} := \frac{1}{2} \left( \underline{\alpha}^{(t)} + \underline{\bar{\alpha}}^{(t)} \right)$ which is symmetric by construction, and has the same value of the dual objective (14) as $\underline{\alpha}^{(t)}$ (as it follows proceeding likewise in the proof of Lemma 1).

## 6. Numerical results

In a first set of numerical simulations, we tested the obtained theoretical results solving the dual problem (13) using LIBSVM, starting from a set of artificial data[10]. For what concerns real-world applications of such properties, we refer, e.g., to our recent work (Dardard et al., 2016) for an application with real motion capture data. Other real-world applications are mentioned in Section 8.

In more details, to obtain the first set of numerical results, we generated the two following scenarios, respectively for the symmetric and antisymmetric case. In both cases, we considered a small number of training examples, just in order to report in the successive tables the optimal dual variables associated with all the support vectors (i.e., in the context of the paper, the ordered pairs of training examples associated with non-zero optimal dual variables), keeping at the same time the size of the tables small.

(a) In the first scenario (with symmetric labels), we randomly generated, with 8 independent draws, an 8-dimensional feature vector of the form

$$\underline{X}_{ab} = (\underline{x}_a, \underline{x}_{ab}, \underline{\tilde{x}}_{ab}, \underline{x}_b) \in \mathbb{R}^8, \tag{35}$$

where each subvector had 2 components, and all such components were sampled independently according to a normal distribution with mean 1 and standard deviation 1, then we attributed the label 1 to these feature vectors. Then, we repeated the procedure above generating other 8 feature vectors of the form (35), using this time a normal distribution with mean $-1$ and standard deviation 1, then we attributed the label $-1$ to these feature vectors. Finally, starting from the 16 feature vectors generated above, we generated other 16 feature vectors of the form

$$\underline{X}_{ba} = (\underline{x}_b, \underline{x}_{ba}, \underline{\tilde{x}}_{ba}, \underline{x}_a) = (\underline{x}_b, \underline{x}_{ab}, -\underline{\tilde{x}}_{ab}, \underline{x}_a) \in \mathbb{R}^8, \tag{36}$$

---

10. Notice that the fact of using artificial data here is not a severe limitation, since the goal of this first numerical study is not to compare the classification performance of one proposed machine learning method with other ones, but to investigate the symmetry/antisymmetry properties of the optimal classifier under suitable conditions on the training set and on the kernel. For the same reason, for this first set of simulations, even the relatively small cardinality of the training set and the absence of an explicit test set are not an issue. Nevertheless, for a more realistic example, see the results of the second set of simulations reported later in this section, where a larger training set was considered, together with a detailed evaluation of the generalized performance through both validation and test sets.

then we attributed to each of them the same label of the associated feature vector of the form (35).

(b) In the second scenario (with antisymmetric labels), we randomly generated, with 16 independent draws, an 8-dimensional feature vector of the form (35) where each subvector had 2 components, and all components were sampled independently according to a normal distribution with mean 1 and standard deviation 1, then we attributed the label 1 to these feature vectors. Starting from these feature vectors, we generated other 16 feature vectors of the form (36), then we attributed the label $-1$ to them.

In both cases, we solved the dual problem (13) using LIBSVM, with the choice $C = 1$ for the penalty parameter, and using, respectively, the linear kernel, the homogeneous polynomial kernel of order $m = 3$, the Gaussian kernel with $\sigma = 1$ (which are all order-invariant kernels, as shown in Section 3), a kernel $K_P$ which is not order-invariant[11] (namely, the one defined as

$$K_P\left(\underline{X}_{ab}, \underline{X}_{cd}\right) := \underline{X}_{ab} P \underline{X}'_{cd}, \tag{37}$$

where $P$ is a $8 \times 8$ symmetric positive definite tridiagonal matrix[12] in which all the elements in the main diagonal are equal to 1, and all the other non-zero elements are equal to 0.2), and an order-invariant kernel generated from $K_P$, namely,

$$K_P^o\left(\underline{X}_{ab}, \underline{X}_{cd}\right) := \frac{1}{2}\left(K_P\left(\underline{X}_{ab}, \underline{X}_{cd}\right) + K_P\left(\underline{X}_{ba}, \underline{X}_{dc}\right)\right) = \frac{1}{2}\left(\underline{X}_{ab} P \underline{X}'_{cd} + \underline{X}_{ba} P \underline{X}'_{dc}\right). \tag{38}$$

We also chose a sufficiently small stopping tolerance parameter (i.e., $10^{-6}$) in LIBSVM. Tables 1 and 2 show, respectively, for the first scenario and for the second scenario, the indices of the support vectors and the corresponding support vector coefficients, defined as the products of the labels of the support vectors with their associated optimal dual variables. The results show that, with all the four order-invariant kernels above, the SMO algorithm implemented in LIBSVM was able to find, in both scenarios, symmetric optimal dual variables[13], which is in accordance with Lemma 1, and that the corresponding support vector coefficients were, respectively, symmetric/antisymmetric. Moreover, for the antisymmetric case, the obtained bias was negligibly small (ranging from the order $10^{-7}$ to the order $10^{-5}$, for the three kernels), which is in accordance with the choice of the bias considered in the proof of Theorem 1). Hence, due to the order-invariance of the kernels, the properties (18) and (19) were satisfied, respectively, which is in accordance with Theorem 1. When the kernel defined by formula (37) was used, the optimal dual variables were not symmetric, which was also expected since such a kernel is not order-invariant, hence Lemma 1 could not be applied. Moreover, although it is not reported in the table, we have also verified that in general, for such a kernel, the properties (18) and (19) were not satisfied. Moreover, for this kernel, when the antisymmetric case was considered, the obtained bias was not negligible (about 0.07, between three and five orders larger than for the other four kernels).

---

11. Indeed, $(1, 1, 1, 1, 1, 1, 1, 1)P(1, 1, 1, 1, 1, 1, 1, 1)' = 15 \neq (1, 1, 1, 1, -1, -1, 1, 1)P(1, 1, 1, 1, -1, -1, 1, 1)' = 11$, which shows that the kernel $K_P$ defined in (37) is not order-invariant.

12. A sufficient condition for the positive definiteness of a symmetric tridiagonal matrix $A \in \mathbb{R}^n \times \mathbb{R}^n$ is its strict dominance, i.e., the property $|A_{ii}| > \sum_{j \neq i, j=1}^n |A_{ij}|$, for all $i = 1, \ldots, n$. This sufficient condition follows from Gersghorin's circle theorem (Dym, 2007, Section 7.2), which states that all the eigenvalues of $A$ belong, in the complex plane, to at least one of the Gersghorin circles $G_i$ (for $i = 1, \ldots, n$), whose centers and radii are defined, respectively, by $A_{ii}$ and $\sum_{j \neq i, j=1}^n |A_{ij}|$.

13. One can notice that, due to the small dimension - with respect to the number of training examples in the two scenarios - of the feature space associated with the linear kernel, the corresponding instance of the dual optimization problem (13) admits in general an infinite number of optimal solutions. In particular, it admits optimal solutions with asymmetric optimal dual variables. In such a case, one can still obtain an optimal solution with symmetric optimal dual variables either applying Theorem 3 (b), or following Remark 6. When there is uniqueness of the optimal solution to the dual optimization problem (13), however, there is symmetry of the optimal dual variables, as it follows from Lemma 1.

| (SV index) | SV coefficient | (SV index) | SV coefficient |
|---|---|---|---|
| (3) 0.2019 | | (19) 0.2019 | |
| (5) 0.0084 | | (21) 0.0084 | |
| (12) -0.2102 | | (28) -0.2102 | |

(a) Linear kernel

| (SV index) | SV coefficient | (SV index) | SV coefficient |
|---|---|---|---|
| (1) 0.0008 | | (17) 0.0008 | |
| (3) 0.0035 | | (19) 0.0035 | |
| (5) 0.0009 | | (21) 0.0009 | |
| (12) -0.0051 | | (28) -0.0051 | |

(b) Polynomial kernel

| (SV index) | SV coefficient | (SV index) | SV coefficient | (SV index) | SV coefficient | (SV index) | SV coefficient |
|---|---|---|---|---|---|---|---|
| (1) 1.0000 | | (17) 1.0000 | | (9) -0.6858 | | (25) -0.6858 | |
| (2) 0.8219 | | (18) 0.8219 | | (10) -0.9137 | | (26) -0.9137 | |
| (3) 0.8276 | | (19) 0.8276 | | (11) -0.8008 | | (27) -0.8008 | |
| (4) 0.9279 | | (20) 0.9279 | | (12) -0.8155 | | (28) -0.8155 | |
| (5) 0.6463 | | (21) 0.6463 | | (13) -0.8812 | | (29) -0.8812 | |
| (6) 0.9221 | | (22) 0.9221 | | (14) -0.8275 | | (30) -0.8275 | |
| (7) 0.7499 | | (23) 0.7499 | | (15) -0.8575 | | (31) -0.8575 | |
| (8) 0.8091 | | (24) 0.8091 | | (16) -0.8778 | | (32) -0.8778 | |

(c) Gaussian kernel

| (SV index) | SV coefficient | (SV index) | SV coefficient |
|---|---|---|---|
| (1) 0.0596 | | (17) 0.0289 | |
| (3) 0.0922 | | (19) 0.1257 | |
| (12) -0.0779 | | (28) -0.2285 | |

(d) Kernel defined by formula (37)

| (SV index) | SV coefficient | (SV index) | SV coefficient |
|---|---|---|---|
| (1) 0.0202 | | (17) 0.0202 | |
| (3) 0.1164 | | (19) 0.1164 | |
| (12) -0.1366 | | (28) -0.1366 | |

(e) Kernel defined by formula (38)

Table 1: Indices of the support vectors (inside the parenthesis) and corresponding support vector coefficients for the first scenario with symmetric labels. The support vectors with indices between 1 and 16 correspond with feature vectors of the form $\underline{X}_{ab} = (\underline{x}_a, \underline{x}_{ab}, \underline{\tilde{x}}_{ab}, \underline{x}_b)$, whereas the support vectors with indices between 17 and 32 correspond with feature vectors of the form $\underline{X}_{ba} = (\underline{x}_b, \underline{x}_{ba}, \underline{\tilde{x}}_{ba}, \underline{x}_a) = (\underline{x}_b, \underline{x}_{ab}, -\underline{\tilde{x}}_{ab}, \underline{x}_a)$, obtained by switching $a$ and $b$ in the first 16 vectors.

To investigate the potential benefit of using an order-invariant kernel when the model generating the data labels respects, respectively, the symmetry/antisymmetry constraint, we also performed a second set of simulations, considering two additional scenarios, in which two pairwise kernels were used, only one of which was order-invariant. To simulate a sufficiently complex and possibly realistic situation, we considered two models that generated symmetric/antisymmetric labels for each pair of objects by applying a threshold to the sum of several features, related to each of the individual and group features of the pair through a low-degree polynomial, with the coefficients of each polynomial unknown to the learning machine. In the third scenario, the labels were generated according to a symmetric label-generation model, while in fourth scenario, an antisymmetric label-generation model was used. For a fair comparison, in order to investigate the effect on the test-set classification accuracy of the presence/absence of the order-invariance property of the pairwise kernel, in these two additional

| (SV index) | SV coefficient |
|---|---|
| (7) 0.0819 | (23) -0.0819 |
| (8) 1.0000 | (24) -1.0000 |
| (13) 0.1900 | (29) -0.1900 |
| (14) 1.0000 | (30) -1.0000 |
| (15) 0.1922 | (31) -0.1922 |
| (16) 0.7334 | (32) -0.7334 |

(a) Linear kernel

| (SV index) | SV coefficient |
|---|---|
| (2) 0.0151 | (18) -0.0151 |
| (4) 0.0005 | (20) -0.0005 |
| (6) 0.0003 | (22) -0.0003 |
| (8) 0.0003 | (24) -0.0003 |
| (9) 0.0024 | (25) -0.0024 |
| (12) 0.0007 | (28) -0.0007 |
| (14) 0.0019 | (30) -0.0019 |
| (15) 0.0002 | (31) -0.0002 |
| (16) 0.0008 | (32) -0.0008 |

(b) Polynomial kernel

| (SV index) | SV coefficient | | |
|---|---|---|---|
| (1) 0.8687 | (17) -0.8687 | (9) 0.9935 | (25) -0.9935 |
| (2) 1.0000 | (18) -1.0000 | (10) 0.9367 | (26) -0.9367 |
| (3) 0.9328 | (19) -0.9328 | (11) 0.8875 | (27) -0.8875 |
| (4) 1.0000 | (20) -1.0000 | (12) 0.8147 | (28) -0.8147 |
| (5) 0.9712 | (21) -0.9712 | (13) 0.9796 | (29) -0.9796 |
| (6) 1.0000 | (22) -1.0000 | (14) 1.0000 | (30) -1.0000 |
| (7) 0.9505 | (23) -0.9505 | (15) 0.9659 | (31) -0.9659 |
| (8) 0.8898 | (24) -0.8898 | (16) 1.0000 | (32) -1.0000 |

(c) Gaussian kernel

| (SV index) | SV coefficient |
|---|---|
| (2) 0.1415 | (24) -1.0000 |
| (7) 0.3613 | (29) -0.2887 |
| (8) 1.0000 | (30) -1.0000 |
| (13) 0.0178 | (31) -0.2320 |
| (14) 1.0000 | (32) -1.0000 |
| (16) 1.0000 | |

(d) Kernel defined by formula (37)

| (SV index) | SV coefficient |
|---|---|
| (7) 0.0217 | (23) -0.0217 |
| (8) 1.0000 | (24) -1.0000 |
| (13) 0.1448 | (29) -0.1448 |
| (14) 1.0000 | (30) -1.0000 |
| (15) 0.1018 | (31) -0.1018 |
| (16) 0.7941 | (32) -0.7941 |

(e) Kernel defined by formula (38)

Table 2: Likewise Table 1, but for the second scenario with antisymmetric labels.

scenarios we only considered, for several realizations of a randomly chosen[14] symmetric and positive semi-definite matrix $P$, kernels of the form $K_P$ (which is not order-invariant with probability 1) and $K_P^o$ (which is always order-invariant).

The models generating the data labels for the two scenarios were constructed as follows. First, two vectors $\underline{c}_1, \underline{c}_2 \in \mathbb{R}^2$ were randomly generated according to a uniform distribution in $\left[-\frac{1}{10}, \frac{1}{10}\right]^2$. Then, two univariate polynomials $p_1$ and $p_2$ of degree 3 were constructed, with coefficients of odd order generated independently according to a uniform distribution in $\left[-\frac{1}{10}, \frac{1}{10}\right]$. All the other coefficients of the polynomials were set to 0. Then, each ordered pair $(a, b)$ of objects was associated with 2-dimensional vectors $\underline{x}_a$ and $\underline{x}_b$ of individual features (each components of which was sampled independently according to a normal distribution with mean 0 and standard deviation $\frac{1}{25}$), while the group feature vectors were defined as[15] $\underline{x}_{ab} = \underline{x}_a + \underline{x}_b$, and $\underline{\tilde{x}}_{ab} = \underline{x}_b - \underline{x}_a$. In the third scenario, the labels were generated according to the following rule (unknown to the learning machine):

$$y_{ab} = \text{sgn}\left(p_1\left(\langle \underline{c}_1, \underline{x}_a \rangle_{\mathbb{R}^2}\right) + p_1\left(\langle \underline{c}_1, \underline{x}_b \rangle_{\mathbb{R}^2}\right) + p_2\left(\langle \underline{c}_2, \underline{x}_{ab} \rangle_{\mathbb{R}^2}\right)\right). \tag{39}$$

---

14. More precisely, first we generated a random square matrix $M \in \mathbb{R}^{8 \times 8}$ (with entries chosen independently according to a uniform distribution in $\left[-\frac{1}{10}, \frac{1}{10}\right]$), then we set $P = M'M$.

Similarly, in the fourth scenario, they were generated as

$$y_{ab} = \text{sgn}\left(p_1\left(\langle \underline{c}_1, \underline{x}_a \rangle_{\mathbb{R}^2}\right) - p_1\left(\langle \underline{c}_1, \underline{x}_b \rangle_{\mathbb{R}^2}\right) + p_2\left(\langle \underline{c}_2, \underline{\tilde{x}}_{ab} \rangle_{\mathbb{R}^2}\right)\right). \tag{40}$$

The constructions above were motivated by the fact that they guarantee the generation of symmetric and antisymmetric labels, respectively. Moreover, for the third scenario, they provide the same average number of positive/negative labels (balancedness of the two classes), whereas for the fourth scenario, the same number of positive/negative labels is guaranteed, since $y_{ba} = -y_{ab}$. After some trials with other choice of the parameters, these were fixed to the values reported above, in order to make the average classification problem sufficiently difficult, to minimize the risk of obtaining a classification accuracy near 100% for both kernels $K_P$ and $K_P^o$.

In order to do training and model selection, we first generated 25 objects for both scenarios. Then, for each of the two kernels $K_P$ and $K_P^o$, the choice of the SVM parameter $C$ was obtained through the following ad-hoc 5-fold cross-validation (with $C$ ranging from $5^{-3}$ to $5^3$, assuming 7 logarithmically spaced values), to exclude, for each fold, any overlap between its training and validation sets. More precisely, in each fold, 5 of the 25 objects were discarded, and used to construct the $4 \cdot 5 = 20$ ordered pairs needed for the validation in that fold. The remaining 20 objects of the fold were used to construct the $19 \cdot 20 = 380$ pairs in its training set. Then, for each of the two kernels $K_P$ and $K_P^o$, the best value of $C$ resulting from the cross-validation was used to re-train the SVM on a training set made of all the $24 \cdot 25 = 600$ pairs of original objects. The final classifiers were then tested on all the $99 \cdot 100 = 9900$ pairs of other 100 objects, which were drawn independently from the same probability distribution as the first 25 objects. The whole procedure (which used for training all the available ordered pairs of objects, likewise in the simulations performed in (Brunner et al., 2012)) was repeated 20 times, each time considering different realizations of $P, \underline{c}_1, \underline{c}_2, p_1, p_2$, and of the training/test objects. The results are reported in Table 3, respectively, and show that, in almost all cases, the choice of an order-invariant kernel improved the test-set classification accuracy[16] of the trained classifier. Indeed, for what concerns the data reported in the first subtable, a better performance was obtained by the order-invariant kernel in 19 cases over 20, while, referring to the data of the second subtable, this occurred in 18 cases over 20. Moreover, for both subtables, the difference in performance was statistically significant, according to a two-sided Wilcoxon signed-rank test (Demšar, 2006) at confidence level $\alpha = 0.05$, which provided the following $p$-values: $p = 0.0001 < \alpha$ for the data reported in the first subtable, and $p = 0.0015 < \alpha$ for the ones of the second subtable.

## 7. Extensions

In the following, we discuss some extensions of the results obtained in the paper.

---

15. For simplicity, in these scenarios, the group features were generated in a very simple way, using the individual features of the objects of the pair (however, see Remark 1 for more complex situations).

16. In this case, due to the balancedness of the two classes, the test-set classification accuracy is a good criterion to evaluate the performance of the classifiers. Otherwise, in other cases, other criteria such as the Area under the Receiving Operating Characteristic Curve (AUC) can be used (Fawcett, 2006).

| | (Repetition number) Classification accuracy (%) | | | | | | | | |
|---|---|---|---|---|---|---|---|---|---|
| $K_P$ | (1) 92.5 | (2) 91.3 | (3) 60.5 | (4) 96.0 | (5) 91.3 | (6) 95.6 | (7) 78.8 | (8) 83.4 | (9) 91.9 | (10) 84.8 |
| | (11) 86.1 | (12) 98.8 | (13) 94.9 | (14) 93.9 | (15) 83.9 | (16) 93.4 | (17) 96.2 | (18) 85.3 | (19) 94.2 | (20) 96.8 |
| $K_P^o$ | (1) 99.1 | (2) 93.6 | (3) 83.7 | (4) 96.0 | (5) 94.6 | (6) 95.8 | (7) 92.1 | (8) 93.1 | (9) 92.5 | (10) 97.1 |
| | (11) 86.4 | (12) 98.7 | (13) 95.7 | (14) 96.5 | (15) 94.4 | (16) 95.7 | (17) 97.8 | (18) 95.2 | (19) 94.3 | (20) 98.0 |

(a) Third scenario

| | (Repetition number) Classification accuracy (%) | | | | | | | | |
|---|---|---|---|---|---|---|---|---|---|
| $K_P$ | (1) 88.7 | (2) 91.3 | (3) 97.0 | (4) 96.6 | (5) 91.9 | (6) 93.3 | (7) 94.1 | (8) 95.4 | (9) 91.1 | (10) 90.2 |
| | (11) 92.7 | (12) 91.8 | (13) 96.8 | (14) 87.2 | (15) 96.0 | (16) 93.9 | (17) 82.6 | (18) 92.1 | (19) 92.3 | (20) 94.3 |
| $K_P^o$ | (1) 91.4 | (2) 93.6 | (3) 97.1 | (4) 98.6 | (5) 93.0 | (6) 93.8 | (7) 98.2 | (8) 97.0 | (9) 95.1 | (10) 99.6 |
| | (11) 87.7 | (12) 94.6 | (13) 98.5 | (14) 88.4 | (15) 97.3 | (16) 93.7 | (17) 89.2 | (18) 94.1 | (19) 95.8 | (20) 98.2 |

(b) Fourth scenario

Table 3: Classification accuracy (in percentage) on the test set, for each repetition of the third scenario (symmetric labels) and of the fourth scenario (antisymmetric labels), for each of the two kernels $K_P$ and $K_P^o$.

## 7.1 Extension to support vector regression

Lemma 1 and Theorem 1 in Section 4.1 can be extended to support vector regression with the linear $\varepsilon$-insensitive loss function (described, e.g., in (Shawe-Taylor and Cristianini, 2004, Section 7.3.3)). In that case, for given $C, \varepsilon > 0$, the primal and dual optimization problems have, respectively, the forms

$$\text{minimize}_{\underline{w} \in E, \gamma \in \mathbb{R}, \{\xi_{ab}, \hat{\xi}_{ab} \in \mathbb{R}: (a,b) \in I\}} \quad \frac{1}{2}\|\underline{w}\|_E^2 + C \sum_{(a,b) \in I} \left(\xi_{ab} + \hat{\xi}_{ab}\right),$$

$$\text{s.t.} \quad \left(\langle \underline{w}, \underline{\phi}^o(\underline{X}_{ab})\rangle_E + \gamma\right) - y_{ab} \leq \varepsilon + \xi_{ab}, \quad \forall (a,b) \in I,$$

$$y_{ab} - \left(\langle \underline{w}, \underline{\phi}^o(\underline{X}_{ab})\rangle_E + \gamma\right) \leq \varepsilon + \hat{\xi}_{ab}, \quad \forall (a,b) \in I,$$

$$\xi_{ab}, \hat{\xi}_{ab} \geq 0, \quad \forall (a,b) \in I, \tag{41}$$

where $y_{ab}, y_{ba} \in \mathbb{R}$, and

$$\text{minimize}_{\{\alpha_{ab}, \hat{\alpha}_{ab} \in \mathbb{R}: (a,b) \in I\}} \quad G^r(\underline{\alpha}),$$

$$\text{s.t.} \quad 0 \leq \alpha_{ab}, \hat{\alpha}_{ab} \leq C, \quad \forall (a,b) \in I,$$

$$\sum_{(a,b) \in I} (\alpha_{ab} - \hat{\alpha}_{ab}) = 0, \tag{42}$$

where

$$G^r(\underline{\alpha}) \;:=\; \frac{1}{2} \sum_{(a,b),(c,d) \in I} (\alpha_{ab} - \hat{\alpha}_{ab}) (\alpha_{cd} - \hat{\alpha}_{cd}) K^o_{ab,cd}$$

$$+ \sum_{(a,b) \in I} \varepsilon (\alpha_{ab} + \hat{\alpha}_{ab}) + \sum_{(a,b),(c,d) \in I} (\alpha_{ab} - \hat{\alpha}_{ab}) y_{ab}.$$

In more details, one obtains the following extension of Lemma 1 to the regression case (for both the symmetric/antisymmetric cases, such an extension was not detailed in (Brunner et al., 2012)).

**Lemma 2** *Let $K^o$ be an order-invariant kernel and one of the following conditions hold:*

*(a) for all $(a, b) \in I$, $y_{ab} = y_{ba}$ (symmetry condition);*

*(b) for all $(a, b) \in I$, $y_{ab} = -y_{ba}$ (antisymmetry condition).*

*Then,*

*(a) there exists an optimal solution $\underline{\alpha}^\circ, \hat{\underline{\alpha}}^\circ$ to the dual optimization problem (42) for which*

$$\alpha^\circ_{ab} = \alpha^\circ_{ba}, \hat{\alpha}^\circ_{ab} = \hat{\alpha}^\circ_{ba}, \quad \forall (a, b) \in I, \tag{43}$$

*(b) there exists an optimal solution $\underline{\alpha}^\circ, \hat{\underline{\alpha}}^\circ$ to the dual optimization problem (42) for which*

$$\alpha^\circ_{ab} = \hat{\alpha}^\circ_{ba}, \hat{\alpha}^\circ_{ab} = \alpha^\circ_{ba}, \quad \forall (a, b) \in I. \tag{44}$$

Theorem 1 extends to the regression case with no changes in the statement, apart from the interpretation of $f^\circ$ as a regression function rather than a classification one. We report in the Appendix a sketch of the proofs of such two extensions. Finally, also Theorem 2 in Section 4.2 could be extended in a similar way to the regression case (to save space and avoid introducing new notation, here we do not provide a detailed investigation of this possible extension, which would require to formulate suitable modifications of the optimization problems (20) and (22)).

## 7.2 Extension to $l_1$-soft margin transductive SVMs for binary classification of ordered pairs

Using the notation of Section 4.1, here we assume that the training set $I$ is decomposed as $I := I_L \cup I_U$, where the two sets $I_L$ and $I_U$ are disjoint and refer, respectively, to the supervised and the unsupervised ordered pairs. We also assume that, when $(a, b)$ belongs to $I_L$ (respectively, $I_U$), also $(b, a)$ belongs to $I_L$ (respectively, $I_U$). Then, the $l_1$-soft margin transductive SVM training problem for binary classification (Joachims, 1999), applied here to the case of the classification of ordered pairs using an order-invariant kernel $K^o$, is formulated as the following optimization problem[17]:

$$\text{minimize}_{\underline{w} \in E, \gamma \in \mathbb{R}, \{\xi_{ab} \in \mathbb{R} : (a,b) \in I\}, \{y_{ab} \in \{-1,+1\} : (a,b) \in I_u\}} \frac{1}{2} \|\underline{w}\|^2_E + C \sum_{(a,b) \in I} \xi_{ab},$$

$$\text{s.t.} y_{ab} \left( \left\langle \underline{w}, \underline{\phi}^o \left( \underline{X}_{ab} \right) \right\rangle_E + \gamma \right) \geq 1 - \xi_{ab}, \quad \forall (a, b) \in I,$$

$$\xi_{ab} \geq 0, \quad \forall (a, b) \in I \tag{45}$$

The difference with respect to the $l_1$-soft margin binary SVM classification of ordered pairs with an order-invariant kernel considered in Section 4.1 is the additional presence of unsupervised ordered pairs, and the fact that also their labels are among the optimization variables, which makes (45) a nonconvex optimization problem. However, for every fixed choice of such labels, the problem has exactly the same form as (12). In order to find a local minimum for large-scale instances of the problem above, the following label-switching algorithm was proposed in (Joachims, 1999) (here formulated in the case of ordered pairs, and with an order-invariant kernel). The algorithm first assign temporary labels to all the unsupervised ordered pairs. Then, a suitable label-switching algorithm is used repeatedly to switch, at each iteration, the temporary labels of two selected ordered pairs. The algorithm has been proved in (Joachims, 1999) to converge in a finite number of iterations to a local minimum of (45), and has been extended in (Sindhwani and Keerthi, 2006) to a multiple label-switching algorithm, which has similar performance guarantees.

---

17. For simplicity and uniformity of notation with Section 4.1, here we assume the same penalty parameter $C > 0$ for all the supervised and unsupervised ordered pairs, but the following discussion can be extended to the case of different penalty parameters. Moreover, we have removed from the formulation an equality constraint, which is present in the formulation of the problem given in (Sindhwani and Keerthi, 2006), but absent in the original formulation of (Joachims, 1999) (however, see also Remark 7 for an extension to this case).

In more details, given a positive integer $S$, the multiple label-switching algorithm of (Sindhwani and Keerthi, 2006) and its selection rule at the iteration $t$ work as follows:

1. solve the problem (45) for the current fixed choice $y_{ab}^{(t)}$ of the labels of the unsupervised ordered pairs $(a, b)$, obtaining the weight vector $\underline{w}^{(t)}$ and the bias $\gamma^{(t)}$;

2. identify unsupervised ordered pairs $(a, b)$ with currently active dual variables $\alpha_{ab}$ and currently positive (respectively, negative) labels $y_{ab}^{(t)}$, and insert them in a sorted list $L^+$ (respectively, in a sorted list $L^-$), sorting the elements in nondecreasing (respectively, nonincreasing) order with respect to the corresponding outputs, which are $\langle \underline{w}^{(t)}, \underline{\phi}^o\,(\underline{X}_{ab}) \rangle_E + \gamma^{(t)}$;

3. pick pairs of elements, one from each list and starting from the top of each list, until either a pair is obtained such that the output of the element from $L^+$ is larger than the output of the element from $L^-$, or $S$ pairs have been picked so far;

4. if at least one pair of elements from the two lists has been selected in step 3, switch the labels of each pair, and go to the next iteration $t + 1$ ; otherwise, terminate the algorithm.

Our following result shows that, if properly initialized and in the case of symmetric (respectively, antisymmetric) labels for the supervised ordered pairs, the multiple label-switching algorithm applied with an order-invariant kernel produces, at each iteration, solutions that satisfy the symmetry (respectively, antisymmetry) constraint.

**Theorem 4** *The following hold.*

(a) *Let $K^o$ be an order-invariant kernel and, for all $(a, b) \in I_L$, let $y_{ab} = y_{ba}$, the initial labels of the unsupervised ordered pairs satisfy, for all $(a, b) \in I_U$, $y_{ab}^{(0)} = y_{ba}^{(0)}$, and the multiple label-switching algorithm be applied with $S$ even, to solve the optimization problem (45). Then, at each iteration $t$, it is possible to switch labels of some unsupervised ordered pairs according to the multiple label-switching algorithm in such a way that, for all $(a, b) \in I_U$, $y_{ab}^{(t+1)} = y_{ba}^{(t+1)}$, and the resulting classifier obtained at step 1 of the next iteration $t + 1$ satisfies, for any ordered pair of objects $(c, d)$, the symmetry constraint*

$$f^{(t+1)}(\underline{X}_{cd}) = f^{(t+1)}(\underline{X}_{dc}) \quad \left( = f^{(t+1)}(\mathcal{T}\underline{X}_{cd}) \right). \tag{46}$$

(b) *A similar statement as (a) holds by removing the condition "S even", and replacing $y_{ab} = y_{ba}$, $y_{ab}^{(0)} = y_{ba}^{(0)}$, and $y_{ab}^{(t+1)} = y_{ba}^{(t+1)}$ with $y_{ab} = -y_{ba}$, $y_{ab}^{(0)} = -y_{ba}^{(0)}$, and $y_{ab}^{(t+1)} = -y_{ba}^{(t+1)}$, respectively, and the symmetry constraint (46) with the antisymmetry constraint*

$$f^{(t+1)}(\underline{X}_{cd}) = -f^{(t+1)}(\underline{X}_{dc}) \quad \left( = -f^{(t+1)}(\mathcal{T}\underline{X}_{cd}) \right). \tag{47}$$

**Remark 7** When $r = \frac{1}{2}$, Theorem 4 (b) holds also for the variation of problem (45) that includes the additional equality constraint stating that a fraction $r$ of unsupervised ordered pairs is classified as positive by the trained machine. This is just the additional equality constraint considered in the transductive SVM formulation reported in (Sindhwani and Keerthi, 2006), which was used to obtain the numerical results of (Dardard et al., 2016). Since the kernel used in that paper was the linear kernel (which is order-invariant), and the initial labels were antisymmetric, we conclude that Theorem 4 (b) provides a justification for the observed antisymmetry of the classifier, which was obtained in (Dardard et al., 2016).

### 7.3 Extension to feature vectors associated with multiple pairs of nodes in a graph

As already noticed, the results obtained in the paper have application in the context of the following machine learning problem: given a graph with a possibly large number of arcs, some of which are supervised, identify a possible relationship between a set of features associated with the nodes joined by the same arc (i.e., ordered pair of nodes) and the (possibly binary, for the case of binary classification, otherwise real-valued, for the case of regression) weight of that arc, under the additional prior knowledge provided by the symmetry (or antisymmetry) of the relationship to be discovered. Now, we consider the following extension, which takes more deeply into account the structure of the graph in doing the prediction (another way to do this is reported at the end of this section). Such an extension consists in using as predictors the features associated not only with the nodes joined by the same arc, but also with their neighbors: e.g., in the case of a $h$-regular graph[18] (i.e., a graph in which each node has the same number $h$ of neighbors), and $h \geq 2$, one could use, to predict the weight of the arc $(a, b)$, feature vectors of the following form:

$$\left( \underline{X}_{n_a^{(1)} a}, \ldots, \underline{X}_{n_a^{(h-1)} a}, \underline{X}_{ab}, \underline{X}_{b n_b^{(1)}}, \ldots, \underline{X}_{b n_b^{(h-1)}} \right) \in \mathbb{R}^{(2h-1)n}, \tag{48}$$

where the nodes $n_a^{(1)}, \ldots, n_a^{(h-1)}$ are the neighbors of $a$ (with the exception of $b$), the nodes $n_b^{(1)}, \ldots, n_b^{(h-1)}$ are the neighbors of $b$ (with the exception of $a$), and each subvector $\underline{X}_{n_a^{(l)} a}$, $\underline{X}_{ab}$, $\underline{X}_{b n_b^{(l)}}$ has the form (1): e.g., in the case of the ordered pair $(n_a^{(1)}, a)$, one has $\underline{X}_{n_a^{(1)} a} := \left( \underline{x}_{n_a^{(1)}}, \underline{x}_{n_a^{(1)} a}, \underline{\tilde{x}}_{n_a^{(1)} a}, \underline{x}_a \right)$. We assume that the classification/regression of the arc $(a, b)$ is invariant with respect to any permutation $\sigma_1$ of the set of feature subvectors $\underline{X}_{n_a^{(1)} a}, \ldots, \underline{X}_{n_a^{(h-1)} a}$ (without permuting the other feature subvectors $\underline{X}_{ab}$ and $\underline{X}_{b n_b^{(1)}}, \ldots, \underline{X}_{b n_b^{(h-1)}}$) and to any permutation $\sigma_2$ of the set of feature subvectors $\underline{X}_{b n_b^{(1)}}, \ldots, \underline{X}_{b n_b^{(h-1)}}$ (again, without permuting the other feature subvectors $\underline{X}_{n_a^{(1)} a}, \ldots, \underline{X}_{n_a^{(h-1)} a}$ and $\underline{X}_{ab}$). This is guaranteed by the following choice of the kernel (compare with item (a) in Section 3):

---

18. Here, we consider such an example in order to have a feature vector of fixed dimension, but in principle one could also extend the analysis to feature vectors of different dimensions (see also Remark 8).

(a') *multiple-pairs kernel*: it is a symmetric and positive semi-definite kernel $K^m$, which is defined on $\mathbb{R}^{(2h-1)n} \times \mathbb{R}^{(2h-1)n}$, and satisfies the additional property

$$K^m\left( \left( \underline{X}_{n_a^{(\sigma_1(1))}a}, \ldots, \underline{X}_{n_a^{(\sigma_1(h-1))}a}, \underline{X}_{ab}, \underline{X}_{bn_b^{(\sigma_2(1))}}, \ldots, \underline{X}_{bn_b^{(\sigma_2(h-1))}} \right), \right.$$

$$\left. \left( \underline{X}_{n_c^{(\sigma_3(1))}c}, \ldots, \underline{X}_{n_c^{(\sigma_3(h-1))}c}, \underline{X}_{cd}, \underline{X}_{dn_d^{(\sigma_4(1))}}, \ldots, \underline{X}_{dn_d^{(\sigma_4(h-1))}} \right) \right)$$

$$= K^m\left( \left( \underline{X}_{n_a^{(1)}a}, \ldots, \underline{X}_{n_a^{(h-1)}a}, \underline{X}_{ab}, \underline{X}_{bn_b^{(1)}}, \ldots, \underline{X}_{bn_b^{(h-1)}} \right), \right.$$

$$\left. \left( \underline{X}_{n_c^{(1)}c}, \ldots, \underline{X}_{n_c^{(h-1)}c}, \underline{X}_{cd}, \underline{X}_{dn_d^{(1)}}, \ldots, \underline{X}_{dn_d^{(h-1)}} \right) \right)$$

for any four (possibly coincident) permutations $\sigma_1$, $\sigma_2$, $\sigma_3$, $\sigma_4$ of the set $\{1, \ldots, h-1\}$. (49)

In the following, we investigate conditions under which, in case of a exchange of the order between $a$ and $b$, the classification/regression of the arc $(b,a)$ as a function of the feature vector

$$\left( \underline{X}_{n_b^{(1)}b}, \ldots, \underline{X}_{n_b^{(h-1)}b}, \underline{X}_{ba}, \underline{X}_{an_a^{(1)}}, \ldots, \underline{X}_{an_a^{(h-1)}} \right) \in \mathbb{R}^{(2h-1)n} \tag{50}$$

is the same as the classification/regression of the arc $(a,b)$ as a function of the feature vector (48) in case of a "cohesion" relationship (i.e., in case of a symmetry constraint), otherwise opposite in sign (but with the same absolute value) in case of a "leadership/followership" relationship (i.e., in case of an antisymmetry constraint)[19]. To investigate these issues, we consider the following multiple-pairs extensions of the definitions of balanced, skew-balanced, and order-invariant kernels introduced in Section 3 (see items (b), (c), and (d) in that section):

(b') *multiple-pairs balanced kernel*: it is a multiple-pairs kernel $K^{mb}$ that satisfies the additional property

$$K^{mb}\left( \left( \underline{X}_{n_a^{(1)}a}, \ldots, \underline{X}_{n_a^{(h-1)}a}, \underline{X}_{ab}, \underline{X}_{bn_b^{(1)}}, \ldots, \underline{X}_{bn_b^{(h-1)}} \right), \right.$$

$$\left. \left( \underline{X}_{n_c^{(1)}c}, \ldots, \underline{X}_{n_c^{(h-1)}c}, \underline{X}_{cd}, \underline{X}_{dn_d^{(1)}}, \ldots, \underline{X}_{dn_d^{(h-1)}} \right) \right)$$

$$= K^{mb}\left( \left( \underline{X}_{n_a^{(1)}a}, \ldots, \underline{X}_{n_a^{(h-1)}a}, \underline{X}_{ab}, \underline{X}_{bn_b^{(1)}}, \ldots, \underline{X}_{bn_b^{(h-1)}} \right), \right.$$

$$\left. \left( \underline{X}_{n_d^{(1)}d}, \ldots, \underline{X}_{n_d^{(h-1)}d}, \underline{X}_{dc}, \underline{X}_{cn_c^{(1)}}, \ldots, \underline{X}_{cn_c^{(h-1)}} \right) \right).$$

---

19. See also footnote 23, reported later in Section 8.

(c') *multiple-pairs skew-balanced kernel*: it is a multiple-pairs kernel $K^{ms}$ that satisfies the additional property

$$K^{ms}\Bigg( \left(\underline{X}_{n_a^{(1)}a}, \ldots, \underline{X}_{n_a^{(h-1)}a}, \underline{X}_{ab}, \underline{X}_{bn_b^{(1)}}, \ldots, \underline{X}_{bn_b^{(h-1)}}\right),$$
$$\left(\underline{X}_{n_c^{(1)}c}, \ldots, \underline{X}_{n_c^{(h-1)}c}, \underline{X}_{cd}, \underline{X}_{dn_d^{(1)}}, \ldots, \underline{X}_{dn_d^{(h-1)}}\right) \Bigg)$$
$$= -K^{ms}\Bigg( \left(\underline{X}_{n_a^{(1)}a}, \ldots, \underline{X}_{n_a^{(h-1)}a}, \underline{X}_{ab}, \underline{X}_{bn_b^{(1)}}, \ldots, \underline{X}_{bn_b^{(h-1)}}\right),$$
$$\left(\underline{X}_{n_d^{(1)}d}, \ldots, \underline{X}_{n_d^{(h-1)}d}, \underline{X}_{dc}, \underline{X}_{cn_c^{(1)}}, \ldots, \underline{X}_{cn_c^{(h-1)}}\right) \Bigg);$$

(d') *multiple-pairs order-invariant kernel*: it is a multiple-pairs kernel $K^{mo}$ that satisfies the additional property

$$K^{mo}\Bigg( \left(\underline{X}_{n_a^{(1)}a}, \ldots, \underline{X}_{n_a^{(h-1)}a}, \underline{X}_{ab}, \underline{X}_{bn_b^{(1)}}, \ldots, \underline{X}_{bn_b^{(h-1)}}\right),$$
$$\left(\underline{X}_{n_c^{(1)}c}, \ldots, \underline{X}_{n_c^{(h-1)}c}, \underline{X}_{cd}, \underline{X}_{dn_d^{(1)}}, \ldots, \underline{X}_{dn_d^{(h-1)}}\right) \Bigg)$$
$$= K^{mo}\Bigg( \left(\underline{X}_{n_b^{(1)}b}, \ldots, \underline{X}_{n_b^{(h-1)}b}, \underline{X}_{ba}, \underline{X}_{an_a^{(1)}}, \ldots, \underline{X}_{an_a^{(h-1)}}\right),$$
$$\left(\underline{X}_{n_d^{(1)}d}, \ldots, \underline{X}_{n_d^{(h-1)}d}, \underline{X}_{dc}, \underline{X}_{cn_c^{(1)}}, \ldots, \underline{X}_{cn_c^{(h-1)}}\right) \Bigg). \tag{51}$$

We also assume that, given any training example with a feature vector of the form (48), all the examples with feature vectors generated applying either a permutation of the form $\sigma_1$ or a permutation of the form $\sigma_2$ to its components are presented to the learning machine with its same label. In this way, due to the property (49), all these examples are actually dealt with as a single example by the learning machine. So, presenting the feature vector $(\underline{X}_{n_a^{(1)}a}, \ldots, \underline{X}_{n_a^{(h-1)}a}, \underline{X}_{ab}, \underline{X}_{bn_b^{(1)}}, \ldots, \underline{X}_{bn_b^{(h-1)}})$ and its label as inputs to training is equivalent (up to rescaling the penalty parameter $C$) to presenting them as inputs together with all the $((h-1)!)^2 - 1$ admissible permutations of that feature vector, using its same label (which would be computationally unfeasible for large $h$). As a consequence, it is possible to identify the ordered pair $(a, b)$ with (any of) the feature vector(s) $(\underline{X}_{n_a^{(1)}a}, \ldots, \underline{X}_{n_a^{(h-1)}a}, \underline{X}_{ab}, \underline{X}_{bn_b^{(1)}}, \ldots, \underline{X}_{bn_b^{(h-1)}})$, and the ordered pair $(b, a)$ with (any of) the feature vector(s) $(\underline{X}_{n_b^{(1)}b}, \ldots, \underline{X}_{n_b^{(h-1)}b}, \underline{X}_{ba}, \underline{X}_{an_a^{(1)}}, \ldots, \underline{X}_{an_a^{(h-1)}})$. In this way, assuming that the training set $I$ contains, for each ordered pair $(a, b)$, also the ordered pair $(b, a)$, the primal and dual optimization problems that model the training of an $l_1$-soft margin binary SVM classifier with a multiple-pairs order-invariant kernel $K^{mo}$ are, respectively,

$$\text{minimize}_{\underline{w} \in E, \gamma \in \mathbb{R}, \{\xi_{ab} \in \mathbb{R}: (a,b) \in I\}} \qquad \frac{1}{2}\|\underline{w}\|_E^2 + C \sum_{(a,b) \in I} \xi_{ab},$$
$$\text{s. t.} \qquad y_{ab}\left( \left\langle \underline{w}, \underline{\phi}^{mo}\left(\underline{X}_{n_a^{(1)}a}, \ldots, \underline{X}_{n_a^{(h-1)}a}, \underline{X}_{ab}, \underline{X}_{bn_b^{(1)}}, \ldots, \underline{X}_{bn_b^{(h-1)}}\right)\right\rangle_E + \gamma \right)$$
$$\geq 1 - \xi_{ab}, \ \ \forall (a,b) \in I,$$
$$\xi_{ab} \geq 0, \ \ \forall (a,b) \in I \tag{52}$$

(where $\underline{\phi}^{mo} : \mathbb{R}^{(2h-1)n} \to E$ denotes the mapping associated with the multiple-pairs order-invariant kernel $\tilde{K}^{mo}$), and

$$\begin{aligned} \text{minimize}_{\{\alpha_{ab} \in \mathbb{R}: (a,b) \in I\}} \quad & \tilde{G}(\underline{\alpha}), \\ \text{s.t.} \quad & 0 \le \alpha_{ab} \le C, \quad \forall (a,b) \in I, \\ & \sum_{(a,b) \in I} y_{ab}\alpha_{ab} = 0, \end{aligned} \tag{53}$$

where

$$\tilde{G}(\underline{\alpha}) := \frac{1}{2} \sum_{(a,b),(c,d) \in I} \alpha_{ab}\alpha_{cd}y_{ab}y_{cd}K^{mo}_{ab,cd} - \sum_{(a,b) \in I} \alpha_{ab}, \tag{54}$$

and $K^{mo}_{ab,cd}$ is a shortcut for

$$K^{mo}\bigg( \Big( \underline{X}_{n_a^{(1)}\,a}, \dots, \underline{X}_{n_a^{(h-1)}\,a}, \underline{X}_{a\,b}, \underline{X}_{b\,n_b^{(1)}}, \dots, \underline{X}_{b\,n_b^{(h-1)}} \Big), \\ \Big( \underline{X}_{n_c^{(1)}\,c}, \dots, \underline{X}_{n_c^{(h-1)}\,c}, \underline{X}_{c\,d}, \underline{X}_{d\,n_d^{(1)}}, \dots, \underline{X}_{d\,n_d^{(h-1)}} \Big) \bigg).$$

These optimization problems have exactly the same forms as (12) and (13), respectively, and the kernel $K^{mo}$ inherits the properties of the kernel $K^o$ appearing therein. Hence, the results of Sections 4 and 5 are extended directly to this situation in which the predictors are the features associated not only with the nodes joined by the same arc, but also with their neighbors, simply replacing the kernels $K^b$, $K^s$, and $K^o$ in those sections by $K^{mb}$, $K^{ms}$, and $K^{mo}$ (here, we do not report the precise statements, because they are practically identical to the ones contained in those sections).

**Remark 8** The simplest way to guarantee that the kernel $K^{mo}$ satisfies the property (49) consists in assuming (using an overloaded notation) that it has the functional form

$$\begin{aligned} & K^{mo}\bigg( \Big( \underline{X}_{n_a^{(1)}\,a}, \dots, \underline{X}_{n_a^{(h-1)}\,a}, \underline{X}_{a\,b}, \underline{X}_{b\,n_b^{(1)}}, \dots, \underline{X}_{b\,n_b^{(h-1)}} \Big), \Big( \underline{X}_{n_c^{(1)}\,c}, \dots, \underline{X}_{n_c^{(h-1)}\,c}, \underline{X}_{c\,d}, \underline{X}_{d\,n_d^{(1)}}, \dots, \underline{X}_{d\,n_d^{(h-1)}} \Big) \bigg) \\ & = K^{mo}\bigg( \Big( \frac{1}{h-1}\sum_{l=1}^{h-1} \underline{X}_{n_a^{(l)}\,a}, \underline{X}_{a\,b}, \frac{1}{h-1}\sum_{l=1}^{h-1} \underline{X}_{b\,n_b^{(l)}} \Big), \Big( \frac{1}{h-1}\sum_{l=1}^{h-1} \underline{X}_{n_c^{(l)}\,c}, \underline{X}_{c\,d}, \frac{1}{h-1}\sum_{l=1}^{h-1} \underline{X}_{d\,n_d^{(l)}} \Big) \bigg), \end{aligned} \tag{55}$$

i.e., that it depends on the subvectors $\underline{X}_{n_a^{(l)}\,a}, \underline{X}_{b\,n_b^{(l)}}, \underline{X}_{n_c^{(l)}\,c}, \underline{X}_{d\,n_d^{(l)}}$ only through the respective means of such subvectors[20]. A similar remark holds for the kernels $K^{mb}$ and $K^{ms}$. One can notice that, in these cases, the graph structure is taken into account taking averages of features of neighboring nodes.

**Remark 9** Starting from any kernel of the form $K^{mo}$, one can define kernels of the forms $K^{mb}$ and $K^{ms}$ using constructions similar to the ones provided by formulas (6) and (7). Also the examples of order-invariant kernels presented in Section 4 are extended directly to analogous examples of kernels of the form $K^{mo}$ for which the property (55) holds. For instance, this holds in the case of the (multiple-pairs) linear kernel, defined as

$$K^{mo}_{\text{lin}}\left(\left(\frac{1}{h-1}\sum_{l=1}^{h-1}\underline{X}_{n^{(l)}_a\,a},\underline{X}_{a\,b},\frac{1}{h-1}\sum_{l=1}^{h-1}\underline{X}_{b\,n^{(l)}_b}\right),\left(\frac{1}{h-1}\sum_{l=1}^{h-1}\underline{X}_{n^{(l)}_c\,c},\underline{X}_{c\,d},\frac{1}{h-1}\sum_{l=1}^{h-1}\underline{X}_{d\,n^{(l)}_d}\right)\right)$$
$$:=\frac{1}{(h-1)^2}\langle\sum_{l=1}^{h-1}\underline{X}_{n^{(l)}_a\,a},\sum_{l=1}^{h-1}\underline{X}_{n^{(l)}_c\,c}\rangle_{\mathbb{R}^{n_1}}+\langle\underline{X}_{a\,b},\underline{X}_{c\,d}\rangle_{\mathbb{R}^{n_1}}+\frac{1}{(h-1)^2}\langle\sum_{l=1}^{h-1}\underline{X}_{b\,n^{(l)}_b},\sum_{l=1}^{h-1}\underline{X}_{d\,n^{(l)}_d}\rangle_{\mathbb{R}^{n_1}}.$$

---

20. Another advantage of this assumption is that it allows an immediate extension of the analysis to the case of graphs in which the nodes can possibly have different numbers of neighbors.

### 7.4 Extension to diffusion kernels on an auxiliary graph

We conclude describing the following other extension of our analysis to the problem of arc classification on graphs, taking into account even more deeply the graph structure. The goal is to show how one can construct a diffusion kernel (Kondor and Lafferty, 2002), (Shawe-Taylor and Cristianini, 2004, Section 9.4) on an auxiliary graph, whose nodes are a selection of the directed arcs of the original graph, providing sufficient conditions under which such diffusion kernel is balanced, or skew-balanced (see the next Theorem 5). A numerical comparison with other (unbalanced or not skew-balanced) graph kernels (Vishwanathan et al., 2010) (including other diffusion kernels), and with the multiple-pairs kernels introduced in Section 7.3, is deferred to a future investigation. In the following, likewise in Section 4.1, we assume that a set $I$ of ordered pairs of distinct objects is given, and that, if $(a,b)$ belongs to $I$, also $(b,a)$ belongs to $I$.

The proposed construction of the auxiliary graph proceeds as follows. For each unordered pair of distinct objects $a,b$, we choose one (and only one) of the two possible orders $(a,b)$ and $(b,a)$, and we associate it with a node of the auxiliary graph. Then, we insert an (undirected) edge between two nodes of the auxiliary graph if and only if they share an object (e.g., we insert an edge between two nodes $(a,b)$ and $(a,c)$, but not between two nodes $(a,b)$ and $c,d$). At this point, we choose a pairwise balanced or skew-balanced kernel $K$ to express the base similarity between two nodes of the auxiliary graph that are connected by an edge (i.e., the base similarity between $(a,b)$ and $(a,c)$ is $K_{ab,ac}$), whereas we set to 0 the base similarity of two nodes of the auxiliary graph that are not connected by an edge. Finally, we denote by $\mathcal{K}$ the corresponding matrix of base similarities, which is indexed by pairs of nodes of the auxiliary graph (for simplicity, in the following we also denote by $\mathcal{K}_{ij}$ its elements, i.e., $i$ and $j$ denote two generic nodes of the auxiliary graph). Then, for a nonincreasing sequence of nonnegative real numbers $\lambda_l$ such that the following series (56) converges, we define the (diffusion) similarity matrix induced by the graph structure as

$$\mathcal{K}^{\text{diff}}:=\sum_{l=0}^{+\infty}\lambda_l\mathcal{K}^l. \tag{56}$$

To compute the similarity induced by the graph structure for a different selection of the nodes of the auxiliary graph, in principle one can repeat the construction above, changing the base similarity matrix $\mathcal{K}$, due to the change in the nodes (in practice, this is not needed, due to the next Theorem 5). In this way, the diffusion kernel[21] $K^{\text{diff}}$ expressing the graph-induced similarities of pairs of nodes is completely defined, apart from the similarities induced by the graph structure between elements of the form $(a, b)$ and $(b, a)$. To define them, we set $K^{\text{diff}}_{(a,b),(b,a)} := K^{\text{diff}}_{(a,b),(a,b)}$ in the case of a balanced kernel, and $K^{\text{diff}}_{(a,b),(b,a)} := -K^{\text{diff}}_{(a,b),(a,b)}$ in the case of a skew-balanced kernel.

---

21. Differently from Section 3, this diffusion kernel is defined on a finite domain. Nevertheless, the concepts of balancedness and skew-balancedness are defined in a similar way.

**Theorem 5** *The following hold.*

(a) *If the pairwise kernel $K$ is balanced or skew-balanced, then the matrix $\mathcal{K}^{\text{diff}}$ is well-defined, in the sense that each element $\mathcal{K}^{\text{diff}}_{ij}$ does not depend on the specific construction of the nodes of the auxiliary graph that are different from $i$ and $j$.*

(b) *The kernel $K^{\text{diff}}$ is symmetric positive semi-definite. If $K$ is balanced (respectively, skew-balanced), so is $K^{\text{diff}}$.*

Concluding, Theorem 5 shows how to define balanced and skew-balanced diffusion kernels, starting from balanced and skew-balanced base kernels.

## 8. Conclusions

We have investigated symmetry and antisymmetry properties of the optimal solutions to $l_1$-soft margin binary SVM classifiers under symmetry/antisymmetry conditions on the labels, extending also some results to support vector regression with the linear $\varepsilon$-insensitive loss function. The results show that, with a suitable choice of the kernel, it is possible to impose such symmetry/antisymmetry properties, which, depending on the specific machine learning problem, may form an additional a-priori knowledge. For the symmetric case, taking the hint from the invariance framework of (Király et al., 2014), the first part of the analysis specializes the one made in (Brunner et al., 2012) to a particular transformation of feature vectors associated with a pair of objects, when one exchanges their order. Moreover, we have also detailed an extension to the antisymmetric case, which was not investigated in (Brunner et al., 2012). Additionally, we have also investigated how a classical algorithm from the literature (one version of the SMO algorithm (Keerthi et al., 2001)) is able to generate a sequence of suboptimal solutions having the same symmetry/antisymmetry properties. Up to our knowledge, this kind of investigation - which we have also performed on an algorithm used in the literature to train transductive SVMs - is completely novel. Numerical examples have confirmed the theoretical results. Finally, we have extended such results to the case in which the feature vectors used to classify/regress the arcs are associated with multiple pairs of nodes, instead than only one such pair, and we have also extended the analysis to diffusion kernels on graphs. The results contribute to fill a current gap in the literature about kernel methods, providing an additional theoretical investigation (besides the ones provided in (Herbrich et al., 1998), (Brunner et al., 2012), (Király et al., 2014) and the other references cited in the Introduction) of symmetry and antisymmetry properties of the optimal solutions to machine learning problems modeled by kernel methods.

For what concerns the numerical results, having used artificial data in the simulations is not a severe limitation, since the goal of our investigation in the first two scenarios was not to compare the classification performance of one proposed machine learning method with other ones, but to investigate numerically the symmetry/antisymmetry properties of the optimal classifier, validating the theoretical results. Moreover, in the third and fourth scenarios, we have compared the test-set classification performance obtained by using different kernels and demonstrated the potential advantage of kernels that impose symmetry/antisymmetry properties, when these are satisfied by the model generating the data labels. For what concerns real-world applications of such properties, we refer to (Brunner et al., 2012) and (Herbrich et al., 1998) for the two cases, respectively. In particular, (Brunner et al., 2012, Section 5.2) deals with a face recognition problem for which the symmetry property arises naturally (the goal therein is to recognize whether two face images belong to the same person or not), while (Herbrich et al., 1998) considers the case in which one wants to learn an order among objects through supervised examples, which naturally leads to the antisymmetric property (this learning framework has applications, e.g., in information retrieval, in regression with ordinal response, and in econometric models). The optimization problem considered therein to model learning is a particular instance of the optimization problem (22), when $K^o$ is the linear kernel, and $C = +\infty$. The present paper extends significantly that setting, by allowing for more general kernels, and providing several additional theoretical results related to the antisymmetric case.

We also refer to the recent work (Dardard et al., 2016) for an application with real motion capture data (incidentally, the antisymmetry property empirically observed when solving numerically the binary classification problem studied in (Dardard et al., 2016)[22] was the source of inspiration for the theoretical investigation of the antisymmetry constraint made in the present work[23]). This application (summarized in footnote 23) refers actually to a setting with both supervised and unsupervised examples, and a theoretical explanation of is numerical results has been provided in Section 7.2. Moreover, for the case of symmetric labels, the developments in that section may be applied to an extension of the supervised face recognition application considered in (Brunner et al., 2012, Section 5.2) to a transductive learning setting, in which one would use both supervised and unsupervised (pairs of) examples. This would be useful for cases in which supervision is time-consuming and costly (see, e.g., (Gnecco et al., 2016, in press) for a situation in which this occurs).

Among other possible future developments, we mention: potential extensions to other kernel methods (e.g., Laplacian SVMs (Belkin and Niyogi, 2006), or combinations of soft and hard constraints (Gnecco et al., 2015b)) of the theoretical results about symmetry/antisymmetry of the optimal solutions; the investigation of possible symmetry/antisymmetry preserving properties of other variations of the SMO algorithm used, e.g., for SVM regression problems.

---

22. See, in particular, (Dardard et al., 2016, Section 5) and the comments to Figures 6 and 7 therein.

23. In more details, the application considered in (Dardard et al., 2016) concerns the analysis of leading interactions in a group of individuals (specifically, musicians in a string quartet). The interactions of the individuals are modeled through a weighted directed graph, where the nodes of the graph represent the individuals, whereas the weight of the arc between any two individuals expresses the intensity of a "leadership/followership" relationship between the two individuals. In this specific case, opposite arcs have opposite labels (in case of a "cohesion" relationship, the same labels would have been used, instead). The feature vector - which is associated with each pair of individuals - is constructed starting from motion capture data associated with the movement of parts of the body of each individual, and inserting in such vector both individual features and group features, obtained through several time-series analysis. After exchanging the order of the two individuals of the pair, one obtains a new feature vector, in which the individual features are only permuted, whereas some of the group features do not change at all (i.e., the "synchronicity" and "level of attention" features defined in Section 3.3 of that paper), others change only in sign (i.e., the "delay" features defined in the same section). In the specific case considered in (Dardard et al., 2016), the weigths of the ordered arcs are obtained by training, as a binary classifier, a transductive SVM with linear kernel (Sindhwani and Keerthi, 2006), whose primal problem has some similarities with the $l_1$-soft margin binary SVM classifier. In the present work, we have focused on the case of the $l_1$-soft margin binary SVM classifier, which is more commonly used in applications (however, an extension of our analysis to transductive SVMs is given in Section 7.2).

## Appendix: Proofs

**Proof of Lemma 1.** The idea of the proof is similar to the one of (Brunner et al., 2012, Lemma 1), with some nontrivial changes in the antisymmetric case. We report here only the one of case (b), since the one of case (a) follows directly by specializing (Brunner et al., 2012, Lemma 1) to the definition of order-invariant kernel reported in this paper (which includes the feature-vector transformation associated with the operator $\mathcal{T}$).

Given any optimal solution $\underline{\alpha}^\star$ to the dual optimization problem (13) under condition (b), one constructs another solution $\underline{\bar{\alpha}}$ defined by $\bar{\alpha}_{ab} := \alpha_{ba}^\star$, for all $(a, b) \in I$, which is feasible for the dual optimization problem (13) since $0 \leq \bar{\alpha}_{ab} \leq C$ for all $(a, b) \in I$, and

$$\sum_{(a,b) \in I} y_{ab} \bar{\alpha}_{ab} = \sum_{(a,b) \in I} y_{ab} \alpha_{ba}^\star = - \sum_{(a,b) \in I} y_{ba} \alpha_{ba}^\star = - \sum_{(a,b) \in I} y_{ab} \alpha_{ab}^\star = 0,$$

where we have exploited the fact that, for each $(a, b) \in I$, the set $I$ contains also the element $(b, a)$. Using also formula (14) and the definition of order-invariant kernel, one gets

$$
\begin{aligned}
2G(\underline{\bar{\alpha}}) &= \sum_{(a,b),(c,d) \in I} \bar{\alpha}_{ab} \bar{\alpha}_{cd} y_{ab} y_{cd} K_{ab,cd}^o - 2 \sum_{(a,b) \in I} \bar{\alpha}_{ab} \\
&= \sum_{(a,b),(c,d) \in I} \alpha_{ba}^\star \alpha_{dc}^\star (-y_{ba})(-y_{dc}) K_{ba,dc}^o - 2 \sum_{(a,b) \in I} \alpha_{ba}^\star \\
&= \sum_{(a,b),(c,d) \in I} \alpha_{ab}^\star \alpha_{cd}^\star y_{ab} y_{cd} K_{ab,cd}^o - 2 \sum_{(a,b) \in I} \alpha_{ab}^\star = 2G(\underline{\alpha}^\star),
\end{aligned}
$$

hence, also $\underline{\bar{\alpha}}$ is optimal for the dual optimization problem (13). Finally, by convexity of that problem and the optimality of $\underline{\alpha}^\star$ and $\underline{\bar{\alpha}}$, respectively, also the solution $\underline{\alpha}^\circ := \frac{1}{2} (\underline{\alpha}^\star + \underline{\bar{\alpha}})$ (which satisfies the symmetry condition (16)) is feasible, and optimal for (13). ∎

**Proof of Theorem 1.** The proof of part (a) follows by the same arguments provided in the proof of (Brunner et al., 2012, Theorem 2). In the following, we report the proof of part (b), which differs considerably from the one of (Brunner et al., 2012, Theorem 2).

By Lemma 1, there exists an optimal solution $\underline{\alpha}^\circ$ to the dual optimization problem (13) for which (16) holds. This, combined with (15), the antisymmetry of the labels, and the definition of order-invariant kernel, provides, for any ordered pair $(c, d)$,

$$
\begin{aligned}
&\left\langle \underline{w}^\circ, \underline{\phi}^o (\underline{X}_{cd}) \right\rangle_E \\
&= \sum_{(a,b) \in I} \alpha_{ab}^\circ y_{ab} \left\langle \underline{\phi}^o ((\underline{X}_{ab})), \underline{\phi}^o (\underline{X}_{cd}) \right\rangle_E \\
&= \sum_{(a,b) \in I} \alpha_{ab}^\circ y_{ab} K_{ab,cd}^o = - \sum_{(a,b) \in I} \alpha_{ba}^\circ y_{ba} K_{ba,dc}^o = - \sum_{(a,b) \in I} \alpha_{ab}^\circ y_{ab} K_{ab,dc}^o \\
&= - \sum_{(a,b) \in I} \alpha_{ab}^\circ y_{ab} \left\langle \underline{\phi}^o (\underline{X}_{ab}), \underline{\phi}^o (\underline{X}_{dc}) \right\rangle_E \\
&= - \left\langle \underline{w}^\circ, \underline{\phi}^o (\underline{X}_{dc}) \right\rangle_E .
\end{aligned}
\tag{57}
$$

Now, since the kernel $K^o$ is order-invariant and the labels satisfy, for every $(a, b) \in I$, the antisymmetry condition $y_{ab} = -y_{ba}$, given any optimal solution $(\underline{w}^\circ, \gamma^\circ, \{\xi_{ab}^\circ \in \mathbb{R} : (a, b) \in I\})$ to the primal optimization problem (12) in which $\underline{w}^\circ$ has the form (15) with optimal dual variables satisfying the symmetry constraints $\alpha_{ab}^\circ = \alpha_{ba}^\circ$, also $\left(\underline{\bar{w}}, \bar{\gamma}, \{\bar{\xi}_{ab} \in \mathbb{R} : (a,b) \in I\}\right)$ (with $\underline{\bar{w}} := \underline{w}^\circ$, $\bar{\gamma} := -\gamma^\circ$, and $\bar{\xi}_{ab} := \xi_{ba}^\circ$) is a feasible solution of the primal optimization problem (12), and provides the same (optimal) value for its objective function. Hence, the closed interval of optimal values for $\gamma$ in the primal optimization problem (12) is symmetric around the origin, so, there exists[24] an optimal solution to the primal optimization problem (12) for which $\gamma^\circ := 0$. With this choice of $\gamma^\circ$, using (57), one obtains (19). ∎

**Proof of Theorem 2.** The idea of the proof is similar to the one of (Brunner et al., 2012, Theorem 3), with some nontrivial changes in the antisymmetric case. We report here only the one of case (b), which differs from that proof due to the presence of the antisymmetric labels, the absence of the bias $\gamma$ among the optimization variables, and the need to check, in a different way, the feasibility of some solutions constructed in the proof. Again, the proof of case (a) follows directly by specializing (Brunner et al., 2012, Theorem 3) to the definition of order-invariant kernel reported in this paper (which includes the feature-vector transformation associated with the operator $\mathcal{T}$).

First of all, recalling that any optimal weight vector $\underline{w}^\circ$ of the primal optimization problem (12) has the representation (15) and is unique, one obtains that

$$\sum_{(a,b) \in I} \alpha_{ab}^\circ y_{ab} K_{ab,cd}^o = \left\langle \underline{w}^\circ, \underline{\phi}^o\left(\underline{X}_{cd}\right) \right\rangle_E \tag{58}$$

is the same for all the optimal solutions $\underline{\alpha}^\circ$ of the optimization problem (13), even when they are not unique. Similarly, since the optimization problem (22) can be interpreted as the dual of a primal one of the form (12) with $\gamma$ removed from the problem formulation (i.e., set to 0), one can show[25] that any optimal weight vector $\underline{\hat{w}}^\circ$ of such a primal optimization problem has the representation

$$\underline{\hat{w}}^\circ := \sum_{(a,b) \in J} \beta_{ab}^\circ y_{ab} \underline{\phi}^s\left(\underline{X}_{ab}\right), \tag{59}$$

where $\beta^\circ$ is any optimal solution to the optimization problem (22). Moreover, it still follows[26] from the proof of (Burges and Crisp, 2000, Theorem 2) that such an optimal weight vector $\underline{\hat{w}}^\circ$ is unique, hence

$$\sum_{(a,b) \in J} \beta_{ab}^\circ y_{ab} K_{ab,cd}^s = \left\langle \underline{\hat{w}}^\circ, \underline{\phi}^s\left(\underline{X}_{cd}\right) \right\rangle_E \tag{60}$$

is the same for all the optimal solutions $\underline{\beta}^\circ$ of the optimization problem (22), even when they are not unique.

---

24. This part of the proof does not extend to case (a) because, when the kernel $K^o$ is order-invariant and the labels satisfy, for every $(a,b) \in I$, the symmetry condition $y_{ab} = y_{ba}$, there is no guarantee that the set of optimal biases $\gamma$'s contains $\gamma = 0$.

25. This is obtained by setting to $\underline{0}$ the gradient vector (with respect to the weight vector) of the Lagrangian associated with that primal optimization problem.

26. Although (Burges and Crisp, 2000, Theorem 2) refers to a primal optimization problem of the form (12) (hence, containing the bias $\gamma$ as an optimization variable), its proof of uniqueness of the optimal weight vector still applies to the present situation in which $\gamma = 0$, since that proof is based on the strict convexity of the objective of the primal optimization problem with respect to the weight vector, i.e., on the strict convexity of the term $\frac{1}{2}\|\underline{w}\|_E^2$, which holds also in the absence of $\gamma$ as an optimization variable.

Due to Lemma 1 (b), among all the optimal solutions of the optimization problem (13), one can choose one for which $\alpha_{ab}^\circ = \alpha_{ba}^\circ$, for every $(a,b) \in I$. Then, one defines the vector $\underline{\bar{\beta}}$ with components

$$\bar{\beta}_{ab} := \alpha_{ab}^\circ + \alpha_{ba}^\circ, \quad \forall (a,b) \in J, \tag{61}$$

which is a feasible solution for the optimization problem (22). Next, one gets

$$\bar{\beta}_{ab} K_{ab,cd}^s = \frac{\bar{\beta}_{ab}}{2} \left( K_{ab,cd}^o - K_{ba,cd}^o \right) = \frac{\alpha_{ab}^\circ + \alpha_{ba}^\circ}{2} \left( K_{ab,cd}^o - K_{ba,cd}^o \right) = \alpha_{ab}^\circ K_{ab,cd}^o - \alpha_{ba}^\circ K_{ba,cd}^o. \tag{62}$$

This, combined with $y_{ab} = -y_{ba}$, implies $\sum_{(a,b) \in I} \alpha_{ab}^\circ y_{ab} K_{ab,cd}^o = \sum_{(a,b) \in J} \bar{\beta}_{ab} y_{ab} K_{ab,cd}^s$. Then, the remaining of the proof consists in showing that $\underline{\bar{\beta}}$ is also an optimal solution of the optimization problem (22) (which proves the equality between (58) and (60), hence, formula (25)), and that the optimal values of the objectives of the two optimization problems (13) and (22) are the same.

The proof of the optimality of $\underline{\bar{\beta}}$ for the optimization problem (22) proceeds as follows. First of all, by using $y_{ab} = -y_{ba}$, the definitions (23) of $K^s$ and (61) of $\underline{\bar{\beta}}$, the symmetry of both $K^o$ and $K^s$, the skew-balancedness of $K^s$, and formula (62) (reversing, in some cases, the roles of $(a,b)$ and $(c,d)$), one obtains

$$\begin{aligned}
& 2G(\underline{\alpha}^\circ) + 2 \sum_{(a,b) \in I} \alpha_{ab}^\circ \\
= {} & \sum_{(a,b) \in I} \alpha_{ab}^\circ y_{ab} \sum_{(c,d) \in I} y_{cd} \alpha_{cd}^\circ K_{ab,cd}^o \\
= {} & \sum_{(a,b) \in I} \alpha_{ab}^\circ y_{ab} \sum_{(c,d) \in J} y_{cd} \left( \alpha_{cd}^\circ K_{ab,cd}^o - \alpha_{dc}^\circ K_{ab,dc}^o \right) \\
= {} & \sum_{(a,b) \in I} \alpha_{ab}^\circ y_{ab} \sum_{(c,d) \in J} y_{cd} \left( \alpha_{cd}^\circ K_{cd,ab}^o - \alpha_{dc}^\circ K_{dc,ab}^o \right) \\
= {} & \sum_{(a,b) \in I} \alpha_{ab}^\circ y_{ab} \sum_{(c,d) \in J} y_{cd} \bar{\beta}_{cd} K_{cd,ab}^s \\
= {} & \sum_{(a,b) \in J} \alpha_{ab}^\circ y_{ab} \sum_{(c,d) \in J} y_{cd} \bar{\beta}_{cd} K_{ab,cd}^s + \sum_{(a,b) \in J} \alpha_{ba}^\circ y_{ba} \sum_{(c,d) \in J} y_{cd} \bar{\beta}_{cd} K_{ba,cd}^s \\
= {} & \sum_{(a,b) \in J} \alpha_{ab}^\circ y_{ab} \sum_{(c,d) \in J} y_{cd} \bar{\beta}_{cd} K_{ab,cd}^s + \sum_{(a,b) \in J} \alpha_{ab}^\circ (-y_{ab}) \sum_{(c,d) \in J} y_{cd} \bar{\beta}_{cd} (-K_{ab,cd}^s) \\
= {} & 2 \sum_{(a,b) \in J} \alpha_{ab}^\circ y_{ab} \sum_{(c,d) \in J} y_{cd} \bar{\beta}_{cd} K_{ab,cd}^s = \sum_{(a,b) \in J} \bar{\beta}_{ab} y_{ab} \sum_{(c,d) \in J} y_{cd} \bar{\beta}_{cd} K_{ab,cd}^s \\
= {} & 2H^s(\underline{\bar{\beta}}) + 2 \sum_{(a,b) \in J} \bar{\beta}_{ab}.
\end{aligned}$$

Then, by exploiting again the definition of $\underline{\bar{\beta}}$, one obtains $2 \sum_{(a,b) \in I} \alpha_{ab}^\circ = 2 \sum_{(a,b) \in J} \bar{\beta}_{ab}$, hence

$$G(\underline{\alpha}^\circ) = H^s(\underline{\bar{\beta}}). \tag{63}$$

Now, given any optimal solution $\underline{\beta}^\circ$ of the optimization problem (22), one defines the vector $\underline{\bar{\alpha}}$ with components

$$\bar{\alpha}_{ab} := \begin{cases} \frac{\beta_{ab}^\circ}{2}, & \text{if } (a,b) \in J, \\ \frac{\beta_{ba}^\circ}{2}, & \text{if } (b,a) \in J, \end{cases} \tag{64}$$

which is a feasible solution for the optimization problem (13). Indeed, $0 \le \bar{\alpha}_{ab} \le C$ for all $(a,b) \in I$, and the constraint $\sum_{(a,b) \in I} y_{ab} \bar{\alpha}_{ab} = 0$ is automatically satisfied, due to $\bar{\alpha}_{ab} = \bar{\alpha}_{ba}$ and $y_{ab} = -y_{ba}$ for all $(a,b) \in I$. Now, by the definition of order-invariant kernel, and the definition (23) of $K^s$ and its symmetry, one gets $\bar{\alpha}_{ab} K^o_{ab,cd} - \bar{\alpha}_{ba} K^o_{ab,dc} = \frac{\beta^\circ_{ab}}{2} \left( K^o_{ab,cd} - K^o_{ab,dc} \right) = \beta^\circ_{ab} K^s_{ab,cd}$. This, combined with $y_{cd} = -y_{dc}$, provides

$$2H^s(\underline{\beta}^\circ) + 2 \sum_{(a,b) \in J} \beta^\circ_{ab}$$

$$= \sum_{(a,b) \in J} \beta^\circ_{ab} y_{ab} \sum_{(c,d) \in J} \beta^\circ_{cd} y_{cd} K^s_{ab,cd} = \sum_{(a,b) \in J} \beta^\circ_{ab} y_{ab} \sum_{(c,d) \in J} \beta^\circ_{cd} y_{cd} \frac{1}{2} \left( K^o_{ab,cd} - K^o_{ba,cd} \right)$$

$$= \frac{1}{2} \sum_{(a,b) \in J} \beta^\circ_{ab} y_{ab} \sum_{(c,d) \in J} \left( \bar{\alpha}_{cd} y_{cd} \left( K^o_{ab,cd} - K^o_{ba,cd} \right) + \bar{\alpha}_{dc} (-y_{dc}) \left( -K^o_{ab,dc} + K^o_{ba,dc} \right) \right)$$

$$= \frac{1}{2} \sum_{(a,b) \in J} \beta^\circ_{ab} y_{ab} \sum_{(c,d) \in J} \bar{\alpha}_{cd} y_{cd} \left( K^o_{ab,cd} - K^o_{ba,cd} \right).$$

Then, by using the definition (64) of $\underline{\bar{\alpha}}$, one obtains $\beta^\circ_{ab} = \bar{\alpha}_{ab} + \bar{\alpha}_{ba}$ for $(a,b) \in J$, and $\bar{\alpha}_{ab} = \bar{\alpha}_{ba}$. Hence, exploiting also $y_{ab} = -y_{ba}$, one gets

$$2H^s(\underline{\beta}^\circ) + 2 \sum_{(a,b) \in J} \beta^\circ_{ab} = \frac{1}{2} \sum_{(a,b) \in J} \beta^\circ_{ab} y_{ab} \sum_{(c,d) \in I} \bar{\alpha}_{cd} y_{cd} \left( K^o_{ab,cd} - K^o_{ba,cd} \right)$$

$$= \sum_{(a,b) \in I} \bar{\alpha}_{ab} y_{ab} \sum_{(c,d) \in I} \bar{\alpha}_{cd} y_{cd} K^o_{ab,cd}$$

$$= 2G(\underline{\bar{\alpha}}) + 2 \sum_{(a,b) \in I} \bar{\alpha}_{ab}.$$

Finally, by exploiting again the definition of $\underline{\bar{\alpha}}$, one obtains $2 \sum_{(a,b) \in J} \beta^\circ_{ab} = 2 \sum_{(a,b) \in I} \bar{\alpha}_{ab}$, then

$$G(\underline{\bar{\alpha}}) = H^s(\underline{\beta}^\circ). \tag{65}$$

Now, assume the $\bar{\underline{\beta}}$ is not an optimal solution to the optimization problem (22). Hence, one obtains $H^s(\underline{\beta}^\circ) < H^s(\underline{\bar{\beta}})$, and, by using (63) and (65), also $G(\underline{\bar{\alpha}}) < G(\underline{\alpha}^\circ)$, which contradicts the assumed optimality of $\underline{\alpha}^\circ$ for the optimization problem (13). Hence, one concludes that $\underline{\bar{\beta}}$ is an optimal solution to the optimization problem (22). ∎

**Proof of Theorem 3.** (a) The fact that $\underline{\alpha}^{(t)}$ is feasible for the dual optimization problem (13) and satisfies the symmetry condition (29) follows from the feasibility of $\underline{\beta}^{(t)}$ for the optimization problem (20), and the definition (34). The convergence to an optimal solution of the dual problem (13) follows from the convergence of the sequence of the $\underline{\beta}^{(t)}$'s to an optimal solution $\underline{\beta}^\circ$ of the optimization problem (20) (Lin, 2001, 2002), and from the fact that, when $\underline{\beta}^{(t)}$ is replaced by $\underline{\beta}^\circ$, formula (64) provides an optimal solution to the dual problem (13), as already shown[27] in the proof of Theorem 2.

---

27. The proof has been detailed for the case of antisymmetric labels, but it is extended straightforwardly to the symmetric case.

(b) Let $t \geq 1$ and assume that condition (29) is satisfied at the iteration $t - 1$. Then, due to such a condition, together with the order-invariance of $K^o$ and the antisymmetry of the labels, one gets

$$
\begin{aligned}
F\left(\alpha_{ab}^{(t-1)}\right) &= \sum_{(c,d) \in I} y_{ab} y_{cd} K_{ab,cd}^o \alpha_{cd}^{(t-1)} - 1 = \sum_{(d,c) \in I} y_{ab} y_{dc} K_{ab,dc}^o \alpha_{dc}^{(t-1)} - 1 \\
&= \sum_{(d,c) \in I} y_{ab} y_{dc} K_{ba,cd}^o \alpha_{dc}^{(t-1)} - 1 = \sum_{(c,d) \in I} (-y_{ba})(-y_{cd}) K_{ba,cd}^o \alpha_{cd}^{(t-1)} - 1 \\
&= \sum_{(c,d) \in I} y_{ba} y_{cd} K_{ba,cd}^o \alpha_{cd}^{(t-1)} - 1 = F\left(\alpha_{ba}^{(t-1)}\right)
\end{aligned} \tag{66}
$$

for all $(a, b) \in I$. This, combined with the antisymmetry of the labels and the selection rule (26), shows that, at the iteration $t$, it is possible to choose the dual variables $\alpha_{a^{(t,1)}b^{(t,1)}}, \alpha_{a^{(t,2)}b^{(t,2)}}$ according to formula (26) and in such a way that $\left(a^{(t,2)}, b^{(t,2)}\right) = \left(b^{(t,1)}, a^{(t,1)}\right)$. Then, the restricted dual optimization problem to be solved at the iteration $t$ (which is obtained by fixing all the dual variables to their previous values, with the exception of $\alpha_{a^{(t,1)}b^{(t,1)}}$ and $\alpha_{a^{(t,2)}b^{(t,2)}} = \alpha_{b^{(t,1)}a^{(t,1)}}$), is

$$
\begin{aligned}
\text{minimize}_{\alpha_{a^{(t,1)}b^{(t,1)}}, \alpha_{b^{(t,1)}a^{(t,1)}} \in \mathbb{R}} \quad & G^{\text{restricted}}\left(\alpha_{a^{(t,1)}b^{(t,1)}}, \alpha_{b^{(t,1)}a^{(t,1)}}\right), \\
\text{s.\,t.} \quad & 0 \leq \alpha_{a^{(t,1)}b^{(t,1)}}, \alpha_{b^{(t,1)}a^{(t,1)}} \leq C, \\
& \sum_{(a,b) \in I^{(t)}} y_{ab} \alpha_{ab} = c^{(t)},
\end{aligned} \tag{67}
$$

where $I^{(t)} := \left\{ \left(a^{(t,1)}, b^{(t,1)}\right), \left(b^{(t,1)}, a^{(t,1)}\right) \right\}$, $c^{(t)} := -\sum_{(a,b) \in I \setminus I^{(t)}} y_{ab} \alpha_{ab}^{(t-1)}$, and

$$
\begin{aligned}
& G^{\text{restricted}}\left(\alpha_{a^{(t,1)}b^{(t,1)}}, \alpha_{b^{(t,1)}a^{(t,1)}}\right) \\
& := \frac{1}{2} \sum_{(a,b),(c,d) \in I^{(t)}} \alpha_{ab} \alpha_{cd} y_{ab} y_{cd} K_{ab,cd}^o + \frac{1}{2} \sum_{(a,b) \in I^{(t)}, (c,d) \in I \setminus I^{(t)}} \alpha_{ab} \alpha_{cd}^{(t-1)} y_{ab} y_{cd} K_{ab,cd}^o \\
& \quad + \frac{1}{2} \sum_{(a,b) \in I \setminus I^{(t)}, (c,d) \in I^{(t)}} \alpha_{ab}^{(t-1)} \alpha_{cd} y_{ab} y_{cd} K_{ab,cd}^o - \sum_{(a,b) \in I^{(t)}} \alpha_{ab} - \sum_{(a,b) \in I \setminus I^{(t)}} \alpha_{ab}^{(t-1)}.
\end{aligned}
$$

Moreover, $c^{(t)} = 0$, due to the antisymmetry of the labels $y_{ab}$, the symmetry of the $\alpha_{ab}^{(t-1)}$'s, and the definition of the set $I \setminus I^{(t)}$. Hence, due to the resulting constraint $\sum_{(a,b) \in I^{(t)}} y_{ab} \alpha_{ab} = 0$, the antisymmetry of the labels $y_{ab}$, and the definition of the set $I^{(t)}$, the optimization problem (67) has clearly a symmetric optimal solution $\alpha_{a^{(t,1)}b^{(t,1)}}^{\circ} = \alpha_{b^{(t,1)}a^{(t,1)}}^{\circ}$. Since the vector $\underline{\alpha}^{(t)}$ is obtained from $\underline{\alpha}^{(t-1)}$ by the updates

$$
\begin{aligned}
\alpha_{a^{(t,1)}b^{(t,1)}}^{(t)} &:= \alpha_{a^{(t,1)}b^{(t,1)}}^{\circ}, \\
\alpha_{b^{(t,1)}a^{(t,1)}}^{(t)} &:= \alpha_{b^{(t,1)}a^{(t,1)}}^{\circ},
\end{aligned}
$$

we conclude that condition (29) is also satisfied at the iteration $t$. ∎

**Sketch of the proofs of the analogues of Lemma 1 and Theorem 1 for support vector regression with the linear $\varepsilon$-insensitive loss function.** For the case of real-valued labels satisfying $y_{ab} = y_{ba}$, given any optimal solution $(\underline{\alpha}^{\star}, \underline{\hat{\alpha}}^{\star})$ of the optimization problem (42), one can obtain

another optimal solution $\underline{\bar{\alpha}}$ defined by $\bar{\alpha}_{ab} := \alpha_{ba}^\star, \bar{\bar{\alpha}}_{ab} := \hat{\alpha}_{ba}^\star$, for all $(a, b) \in I$, whereas, for the case of real-valued labels satisfying $y_{ab} = -y_{ba}$, one can obtain another optimal solution $\underline{\bar{\alpha}}$ defined by $\bar{\alpha}_{ab} := \hat{\alpha}_{ba}^\star, \bar{\bar{\alpha}}_{ab} := \alpha_{ba}^\star$, for all $(a, b) \in I$. Then, the remainings of the proofs of the analogues of Lemma 1 and Theorem 1 for support vector regression with the linear $\varepsilon$-insensitive loss function proceed as in the Appendix, taking also into account (Burges and Crisp, 2000, Theorem 3), which is the analogue of (Burges and Crisp, 2000, Theorem 2) for this regression context. ∎

**Proof of Theorem 4.** (a). At the iteration $t = 0$ of the multiple label-switching algorithm, the optimization problem to be solved at its step 1 has symmetric labels, and has the same form as the problem (12). Then, step 1 can be performed using the version of the SMO algorithm detailed in Section 5, as in Theorem 3 (a)[28]. As a consequence, the resulting classifier has weight vector $\underline{w}^{(0)}$ and bias $\gamma^{(0)}$ satisfying, for any ordered pair $(c, d)$, the symmetry constraint $f^{(0)}(\underline{X}_{cd}) = f^{(0)}(\underline{X}_{dc})$, and the current dual variables satisfy $\alpha_{ab}^{(0)} = \alpha_{ba}^{(0)}$, for all $(a, b) \in I$. Then, the two sorted lists $L^+$ and $L^-$ at the iteration 0 can be constructed in such a way that each ordered pair in the list (say, $(\hat{a}, \hat{b})$) is immediately followed by the reversed ordered pair $(\hat{b}, \hat{a})$ in the same list (since they have the same current dual variables, labels, and outputs). As a consequence, exploiting also the fact that the number $S$ is even, if some multiple label-switching is performed at the step 4, then both the ordered pairs $(\hat{a}, \hat{b})$ and $(\hat{b}, \hat{a})$ have their labels switched at the next iteration $t + 1 = 1$. Then, the optimization problem to be solved at the step 1 of the iteration 1 (if such an iteration is performed) has again symmetric labels, and has the same form as the problem (12). Hence, one can proceed as at the iteration 0, and the proof is completed using an induction argument.

(b) In this case, at the iteration $t = 0$ of the multiple label-switching algorithm, the optimization problem to be solved at its step 1 has antisymmetric labels, and has the same form as the problem (12). Then, step 1 can be performed using the version of the SMO algorithm detailed in Section 5, this time as in Theorem 3 (b). As a consequence, the resulting classifier has weight vector $\underline{w}^{(0)}$ and bias $\gamma^{(0)}$ satisfying, for any ordered pair $(c, d)$, the antisymmetry constraint $f^{(0)}(\underline{X}_{cd}) = -f^{(0)}(\underline{X}_{dc})$, and the current dual variables satisfy $\alpha_{ab}^{(0)} = \alpha_{ba}^{(0)}$, for all $(a, b) \in I$. Then, the two sorted lists $L^+$ and $L^-$ at the iteration 0 can be constructed in such a way that each ordered pair in the list (say, $(\hat{a}, \hat{b})$) has the reversed ordered pair $(\hat{b}, \hat{a})$ in the same position in the other list (since they have the same current dual variables, but opposite labels and outputs). As a consequence, if some multiple label-switching is performed at the step 4, then both the ordered pairs $(\hat{a}, \hat{b})$ and $(\hat{b}, \hat{a})$ have their labels switched[29] at the next iteration $t + 1 = 1$. Then, the optimization problem to be solved at the step 1 of the iteration 1 (if such an iteration is performed) has again antisymmetric labels, and has the same form as the problem (12). Hence, one can proceed as at the iteration 0, and the proof is completed again using an induction argument. ∎

**Proof of Theorem 5.** (a) Let $h = (\hat{a}, \hat{b})$ be a generic node of the auxiliary graph, different from $i$ and $j$, and suppose that $h$ is redefined in the auxiliary graph as $h = (\hat{b}, \hat{a})$. Then, when the kernel $K$ is balanced, all the elements of the matrix $\mathcal{K}$ (hence, of the matrix $\mathcal{K}^{\mathrm{diff}}$) are unchanged as a consequence of this change in the definition of $h$. Instead, when $K$ is skew-balanced, all the elements of $\mathcal{K}$ with one index equal to $h$ change in sign. However, this does not change the value of $\mathcal{K}_{ij}^{\mathrm{diff}}$, since, for every nonnegative integer $l$, the generic element of $\mathcal{K}_{ij}^l$ is the summation, over all paths of length $l$ joining the nodes $i$ and $j$ of the auxiliary graph, of the products of the $l$ weights $\mathcal{K}_{\hat{i}_h \hat{j}_h}$ encountered along each path (with $\hat{i}_1 = i$, and $\hat{j}_l = j$), and all such paths cannot contain an odd number of times edges joining $h$ with other nodes of the auxiliary graph.

---

28. In this proof, differently from Section 5, we use the superscript $t = 0$ to refer to the solution obtained at the termination of the SMO algorithm, when concluding the step 1 of the multiple label-switching algorithm at its iteration $t = 0$.

29. In this case, the condition "$S$ even" used in part (a) is not needed.

(b) If the kernel $K$ is balanced and the node $h$ which is changed in part (a) of the proof is replaced by $j$, then the matrices $\mathcal{K}$ and $\mathcal{K}^{\mathrm{diff}}$ do not change, hence also the value of $\mathcal{K}_{ij}^{\mathrm{diff}}$ does not change. Instead, if the kernel $K$ is skew-balanced, if $i$ is different from $j$, all the terms $\mathcal{K}_{ij}^l$ change in sign, since the node $j$ is encountered an odd number of times along each path mentioned in the proof of part (a) (the case in which $i$ coincides with $j$ is of not interest, and leads to no contradiction). Now, let us fix a choice for the nodes in the auxiliary graph. Then, using the first $\frac{|I|}{2}$ indices for these nodes, and the last $\frac{|I|}{2}$ for the corresponding nodes associated with the opposite ordered pairs, the first lines in the proof of this part (b) show that the diffusion kernel (matrix) $K^{\mathrm{diff}}$ has the structure

$$K^{\mathrm{diff}} = \begin{pmatrix} \mathcal{K}^{\mathrm{diff}} & \mathcal{K}^{\mathrm{diff}} \\ \mathcal{K}^{\mathrm{diff}} & \mathcal{K}^{\mathrm{diff}} \end{pmatrix} \tag{68}$$

when $K$ is balanced, and

$$K^{\mathrm{diff}} = \begin{pmatrix} \mathcal{K}^{\mathrm{diff}} & -\mathcal{K}^{\mathrm{diff}} \\ -\mathcal{K}^{\mathrm{diff}} & \mathcal{K}^{\mathrm{diff}} \end{pmatrix} \tag{69}$$

when $K$ is skew-balanced. In both cases, the matrix $K^{\mathrm{diff}}$ is symmetric positive semi-definite, because $\mathcal{K}^{\mathrm{diff}}$ is symmetric positive semi-definite by the definition (56) (every matrix $\mathcal{K}^l$ is symmetric positive semi-definite, and the coefficients $\lambda_l$ are nonnegative), and, for any row vectors $\underline{x}, \underline{y} \in \mathbb{R}^{\frac{|I|}{2}}$, one has

$$\begin{pmatrix} \underline{x} & \underline{y} \end{pmatrix} \begin{pmatrix} \mathcal{K}^{\mathrm{diff}} & \mathcal{K}^{\mathrm{diff}} \\ \mathcal{K}^{\mathrm{diff}} & \mathcal{K}^{\mathrm{diff}} \end{pmatrix} \begin{pmatrix} \underline{x}' \\ \underline{y}' \end{pmatrix} = (\underline{x} + \underline{y})\mathcal{K}^{\mathrm{diff}}(\underline{x} + \underline{y})' \geq 0, \tag{70}$$

and

$$\begin{pmatrix} \underline{x} & \underline{y} \end{pmatrix} \begin{pmatrix} \mathcal{K}^{\mathrm{diff}} & -\mathcal{K}^{\mathrm{diff}} \\ -\mathcal{K}^{\mathrm{diff}} & \mathcal{K}^{\mathrm{diff}} \end{pmatrix} \begin{pmatrix} \underline{x}' \\ \underline{y}' \end{pmatrix} = (\underline{x} - \underline{y})\mathcal{K}^{\mathrm{diff}}(\underline{x} - \underline{y})' \geq 0. \tag{71}$$

Finally, the proof of the balancedness (respectively, skew-balancedness) of $K^{\mathrm{diff}}$ follows directly from (68) (respectively, from (69)). ∎